%% file: 00_main.tex
\documentclass[lettersize,journal]{IEEEtran}
\usepackage[colorlinks=true,
            citecolor=blue, linkcolor = blue,
            urlcolor  = blue,
            anchorcolor = blue]{hyperref}

\usepackage{amsmath,amsfonts}
\usepackage{algorithmic}
\usepackage{algorithm}
\usepackage{array}
\usepackage{booktabs}
\usepackage{caption}
\usepackage{cite}
\usepackage{graphicx}
\usepackage{makecell}
\usepackage{orcidlink}
\usepackage{pifont}
\usepackage{subcaption}
\usepackage{stfloats}
\usepackage{textcomp}
\usepackage{url}
\usepackage{verbatim}
\usepackage{xcolor}
\usepackage{mathtools}
\usepackage{lipsum} 
\usepackage{booktabs}
\usepackage{makecell}
\usepackage{color}
\usepackage{enumitem}
\usepackage{fontawesome5}
\usepackage{cleveref} 

\newcounter{customsubparagraph}[subsection]
\renewcommand{\thecustomsubparagraph}{\emph{\roman{customsubparagraph}}.}

\newcommand{\subparagraphnumbered}[1]{%
  \refstepcounter{customsubparagraph} 
  \hspace{2mm} \thecustomsubparagraph\quad #1 
}

\usepackage{xcolor,soul}
\definecolor{myhlcolor}{HTML}{e2eefa}
\DeclareRobustCommand{\hlmycolor}[1]{{\sethlcolor{myhlcolor}\hl{#1}}}

\newcommand{\FY}[1]{f_{Y}^{(#1)}}
\newcommand{\FX}[1]{f_{X}^{(#1)}}
\newcommand{\FYX}[1]{f_{Y|X}^{(#1)}}
\newcommand{\FXY}[1]{f_{X|Y}^{(#1)}}

\begin{document}
\title{Non-IID data in Federated Learning: A Survey with Taxonomy, Metrics, Methods, Frameworks and Future Directions}

\author{Daniel M. Jimenez G.\,\orcidlink{0000-0002-0305-814X}, David Solans\,\orcidlink{0000-0001-6979-9330}, Mikko A. Heikkil{\"a}\,\orcidlink{0000-0001-5753-8643}, Andrea Vitaletti\,\orcidlink{0000-0003-1074-5068}, Nicolas Kourtellis\,\orcidlink{0000-0002-5674-1698} Aris Anagnostopoulos\,\orcidlink{0000-0001-9183-7911}, Ioannis Chatzigiannakis\,\orcidlink{0000-0001-8955-9270},~\IEEEmembership{Senior Member,~IEEE}
\thanks{Manuscript received September XX, 2024; revised September XX, 2024.}}

\markboth{IEEE COMMUNICATION SURVEYS \& TUTORIALS,~Vol.~00, No.~00, September~2024}%
{Shell \MakeLowercase{\textit{et al.}}: A Sample Article Using IEEEtran.cls for IEEE Journals}


\maketitle

\input{Sections/0.Abstract}

\begin{IEEEkeywords}
federated learning, data heterogeneity, non-IID data, distributed learning, partition protocols, non-IID metrics
\end{IEEEkeywords}

\input{Sections/1.Introduction}

\input{Sections/2.FL_basics}

\input{Sections/3.Literature_review}

\input{Sections/4.Taxonomy_non_iid}

\input{Sections/5.Protocols_and_metrics}

\input{Sections/6.solutions}

\input{Sections/7.Frameworks_for_FL}

\input{Sections/8.Lessons}


\input{Sections/9.Future_directions}

\input{Sections/10.Conclusion}

\section{Acknowledgments}
Daniel Mauricio Jimenez G. and Andrea Vitaletti were partially supported by
PNRR351 TECHNOPOLE -- NEXT GEN EU Roma Technopole -- Digital Transition,
FP2 -- Energy transition and digital transition in urban
regeneration and construction and Sapienza Ateneo Research grant
``La disintermediazione della Pubblica Amministrazione: il ruolo della
tecnologia blockchain e le sue implicazioni nei processi e nei ruoli
della PA.'' 
Aris Anagnostopoulos was supported by the ERC
Advanced Grant 788893 AMDROMA, the EC H2020RIA project ``SoBigData++''
(871042), the PNRR MUR project PE0000013-FAIR, the PNRR MUR project
IR0000013-SoBigData.it, and the MUR PRIN project 2022EKNE5K ``Learning in
Markets and Society.''
Ioannis Chatzigiannakis was
supported by  PE07-SERICS (Security and Rights in the Cyberspace) --
European Union Next-Generation-EU-PE0000014 (Piano Nazionale di Ripresa
e Resilienza -- PNRR).
Andrea Vitaletti was
supported by  PE11 - MICS (Made in Italy -- Circular and Sustainable) --
European Union Next-Generation-EU (Piano Nazionale di Ripresa e
Resilienza -- PNRR) and the project SERICS (PE00000014) under the MUR National Recovery and Resilience Plan funded by the European Union - NextGenerationEU.

Telefónica's contribution was partially supported by The Ministry of Economic Affairs and Digital Transformation of Spain and the European UnionNextGenerationEU programme for the "Recovery, Transformation and Resilience Plan" and the "Recovery and Resilience Mechanism" under agreements TSI-063000-2021-142 and TSI-063000-2021-147 (6G-RIEMANN). Additionally, this research has been as well partially supported by funding received from the Smart Networks and Services Joint Undertaking (SNS JU) under the European Union's Horizon Europe research and innovation programme under Grant Agreement No 101096435 (CONFIDENTIAL6G).


\bibliographystyle{IEEEtran}
\bibliography{00_references}

\end{document}

%% file: Sections/0.Abstract.tex
\begin{abstract}

Recent advances in machine learning have highlighted Federated Learning (FL) as a promising approach that enables multiple distributed users (so-called clients) to collectively train ML models without sharing their private data. While this privacy-preserving method shows potential, it struggles when data across clients is not independent and identically distributed (non-IID) data. The latter remains an unsolved challenge that can result in poorer model performance and slower training times. Despite the significance of non-IID data in FL, there is a lack of consensus among researchers about its classification and quantification. This technical survey aims to fill that gap by providing a detailed taxonomy for non-IID data, partition protocols, and metrics to quantify data heterogeneity. Additionally, we describe popular solutions to address non-IID data and standardized frameworks employed in FL with heterogeneous data. Based on our state-of-the-art survey, we present key lessons learned and suggest promising future research directions.

\end{abstract}

%% file: Sections/1.Introduction.tex
\section{Introduction}
Federated Learning (FL)~\cite{mcmahan2017communication,chai2024survey} represents a paradigm shift in the development of machine learning (ML) models, particularly in response to the growing need for data privacy and the distributed nature 
of modern datasets. 
Traditionally, ML  frameworks have been designed on the assumption that data is centralized, residing in a single, unified repository.
This centralized approach allows for directly aggregating data from diverse sources, facilitating the training of models on large, comprehensive datasets.
However, such a methodology introduces several challenges, mainly when dealing with sensitive or proprietary data.
Data centralization can lead to privacy concerns, increased security risks, and logistical difficulties associated with transferring and storing vast amounts of information.

As data becomes increasingly decentralized (a.k.a. federated), residing on edge devices such as smartphones or within isolated institutional silos (so-called clients), limitations of traditional ML approaches such as privacy and efficiency have become more pronounced.
Data privacy regulations, such as the General Data Protection Regulation (GDPR)~\cite{ullagaddi2024gdpr} and the Health Insurance Portability and Accountability Act (HIPAA)~\cite{lincke2024complying}, impose substantial requirements on collecting and processing personal data, making the centralized aggregation of such data impractical or legally infeasible.

FL addresses these concerns by enabling collaborative model training across decentralized data sources without requiring raw data to leave its local environment.
In FL, local models are trained independently on each client, and only the learned model updates (e.g., model gradients or weights) are shared with a central server.
The central server then aggregates these updates, typically through weighted averaging, to create a global model incorporating knowledge from all participating nodes. 
Crucially, no raw data is exchanged, which helps preserve privacy and mitigate the risks associated with data breaches or leakage during transmission.

This decentralized approach is particularly suited to applications where data heterogeneity, privacy, and security are paramount, such as healthcare~\cite{jimenez2023application}, finance~\cite{awosika2024transparency}, and mobile computing~\cite{deng2024fedasa}. 
FL thus provides a robust solution to the challenges posed by the distributed nature of modern data while maintaining the utility of machine learning models at scale.

\hlmycolor{Throughout this paper, we highlight key findings from our survey using light blue background text to facilitate quick identification of significant insights.}

\subsection{Motivation}

In FL research, there is a notable lack of publicly available datasets that are inherently federated and effectively capture the diverse and non-centralized data distributions typical of real-world environments. 
%
Consequently, a widely adopted approach in the literature involves simulating the federated setting by partitioning existing centralized datasets among several synthetic clients. 
This explicit partitioning allows for emulating a decentralized learning environment across multiple clients while retaining the benefits of a controlled experimental scenario.

Typically, for this synthetic data partitioning, centralized datasets such as CIFAR-10~\cite{krizhevsky2009learning}, MNIST~\cite{lecun2010mnist} or FMNIST~\cite{xiao2017fashion} are divided into non-overlapping datasets, which are then assigned to individual clients.
Usually, various types of data skewness, such as label imbalance, feature distribution divergence, and quantity skew, can be systematically generated through specific partition protocols.
We use the term \emph{partition protocol} (a.k.a partition method) to refer to the systematic partitioning functions used to divide a dataset into smaller subsets, leading to a federated data environment. 
These partition protocols are crucial in simulating meaningful experimental scenarios for FL. 
Depending on the protocol, the resulting distributions can either follow an IID (independent and identically distributed) scheme, where each client receives data that mirrors the overall distribution of the dataset, or a non-IID scheme (a.k.a. non-IIDness), where data is distributed in a skewed or heterogeneous manner across clients.

The choice of a partition protocol significantly impacts the performance and convergence of FL algorithms, as non-IID data often presents additional challenges for model generalization and parameter aggregation compared to an IID data setting.
Thus, designing and selecting appropriate partition protocols is critical to evaluating the robustness and effectiveness of FL methods.

\subsection{Previous surveys on non-IID data in FL}
However, despite the critical importance of establishing a solid understanding of the existing types of data skewness in FL and the necessity of using standardized approaches to simulate realistic data partitions, \hlmycolor{few studies have thoroughly addressed the need to create a meaningful and comprehensive categorization of existing data partitioning techniques in FL}.
While we have identified a few surveys that provide foundational insights into the challenges of data heterogeneity, our work advances beyond these contributions by offering an extended taxonomy alongside a more detailed overview of existing protocols and metrics.

We present previous surveys and reviews in Table~\ref{tab:surveys_comparison}, which reveals a clear evolution in the coverage of non-IID data aspects from 2021 to 2024. Earlier surveys from 2021-2022 demonstrate partial or limited coverage across most categories, although they generally include solutions to non-IIDness. Recent surveys from 2024 still maintain significant gaps, particularly in the inclusion of partition protocols, non-IID metrics, modality skew, and standardized frameworks (see Table~\ref{tab:frameworks_compare}). Our work includes complete coverage of all the aspects related to non-IID data and represents a significant advancement towards understanding, quantifying, and tackling non-IIDness in FL. 

\begin{table*}[t!]
  
  \resizebox{\textwidth}{!}{%
    \begin{tabular}{cccccccccc}
      \toprule
      \textbf{\makecell{Survey or \\ Review}} & \textbf{\makecell{Publication \\ Year}} & \textbf{\makecell{non-IID data \\ Taxonomy}} & \textbf{\makecell{Partition  \\ Protocols}} & \textbf{\makecell{non-IID \\ Metrics}} & \textbf{\makecell{Modality \\ Skew}} & \textbf{\makecell{Solutions\\ to non-IIDness}} & \textbf{\makecell{Standardized\\Frameworks}} \\
      \midrule
      \textbf{\cite{mora2024enhancing}} & 2024 & \textcolor{purple}{\faTimesCircle} & \textcolor{purple}{\faTimesCircle} & \textcolor{purple}{\faTimesCircle} & \textcolor{purple}{\faTimesCircle} & \textcolor{teal}{\faCheckCircle} & \textcolor{purple}{\faTimesCircle} \\
      \textbf{\cite{lu2024federated}} & 2024 & \textcolor{purple}{\faTimesCircle} & \textcolor{purple}{\faTimesCircle} & \textcolor{purple}{\faTimesCircle} & \textcolor{orange}{\faMinusCircle} & \textcolor{teal}{\faCheckCircle} & \textcolor{purple}{\faTimesCircle} \\  
      \textbf{\cite{pei2024review}} & 2024 & \textcolor{orange}{\faMinusCircle} & \textcolor{purple}{\faTimesCircle} & \textcolor{purple}{\faTimesCircle} & \textcolor{purple}{\faTimesCircle} & \textcolor{teal}{\faCheckCircle} & \textcolor{purple}{\faTimesCircle} \\       
      \textbf{\cite{chen2024advances}} & 2024 & \textcolor{teal}{\faCheckCircle} & \textcolor{purple}{\faTimesCircle} & \textcolor{purple}{\faTimesCircle} & \textcolor{purple}{\faTimesCircle} & \textcolor{teal}{\faCheckCircle} & \textcolor{purple}{\faTimesCircle} \\      
      \textbf{\cite{criado2022non}} & 2022 & \textcolor{orange}{\faMinusCircle} & \textcolor{purple}{\faTimesCircle} & \textcolor{purple}{\faTimesCircle} & \textcolor{purple}{\faTimesCircle} & \textcolor{orange}{\faMinusCircle} & \textcolor{purple}{\faTimesCircle} \\
      \textbf{\cite{ma2022state}} & 2022 & \textcolor{orange}{\faMinusCircle} & \textcolor{orange}{\faMinusCircle} & \textcolor{purple}{\faTimesCircle} & \textcolor{purple}{\faTimesCircle} & \textcolor{teal}{\faCheckCircle} & \textcolor{purple}{\faTimesCircle} \\
      \textbf{\cite{zhu2021federated}} & 2021 & \textcolor{teal}{\faCheckCircle} & \textcolor{orange}{\faMinusCircle} & \textcolor{purple}{\faTimesCircle} & \textcolor{purple}{\faTimesCircle} & \textcolor{teal}{\faCheckCircle} & \textcolor{orange}{\faMinusCircle} \\
      \textbf{Ours} & 2024 & \textcolor{teal}{\faCheckCircle} & \textcolor{teal}{\faCheckCircle} & \textcolor{teal}{\faCheckCircle} & \textcolor{teal}{\faCheckCircle} & \textcolor{teal}{\faCheckCircle} & \textcolor{teal}{\faCheckCircle} \\      
      \bottomrule
    \end{tabular}%
  }
    \caption{Comparison of previous surveys and reviews for non-IIDness in FL (\textcolor{teal}{\faCheckCircle}: Included, \textcolor{orange}{\faMinusCircle}: Partially included, \textcolor{purple}{\faTimesCircle}: Not included)}
    \label{tab:surveys_comparison}
\end{table*}


\subsection{Contribution}

This work is the first technical survey dedicated to organizing and synthesizing the existing knowledge regarding distribution skewness in FL, providing a comprehensive taxonomy and detailed categorization of data heterogeneity types. We specifically contribute by:

\begin{enumerate}
    \item Introducing a taxonomy that organizes and synthesizes the diverse forms of non-IIDness.
    \item Reviewing state-of-the-art partition protocols for simulating non-IID scenarios in FL.
    \item Methodically analyzing non-IID metrics for quantifying data heterogeneity, fostering standardized evaluations.
    \item Offering good practices to ensure experimental consistency, fair benchmarking, and realistic deployment testing.
\end{enumerate}

Our work bridges gaps in existing surveys, serving as a critical reference for researchers and practitioners in FL.

\subsection{Paper structure}
The rest of this survey is organized as follows:
Section~\ref{sec:FL-basics} provides an overview of the basic notions of FL. %
Section~\ref{sec:SLR} describes the approach utilized to retrieve the most relevant papers this survey.
Section~\ref{sec:data-heterogeneity} categorizes the various types of data skewness considered in the state of the art.
Section~\ref{sec:protocols-and-metrics} overviews the landscape of partition protocols and heterogeneity metrics utilized in previous research.
Section~\ref{sec:non-iid-solutions} reports existing state-of-the-art solutions to tackle non-IIDness in FL.
Section~\ref{sec:frameworks} collects information about existing standardized frameworks for FL and their capacities to deal with heterogeneous data.
Section~\ref{sec:lessons} compiles a collection of lessons learned and a list of good practices based on the knowledge acquired during the execution of this work. 
Section{~\ref{sec:future} summarizes the future work vision from the various research efforts analyzed.
Finally, Section~\ref{sec:conclusions} synthesizes the key findings from our survey and overviews the potential directions for future work in the field. Fig.~\ref{fig:outline} illustrates the organizational structure of our survey.

\begin{figure*}[ht]
  \centering
  \includegraphics[width=\linewidth]{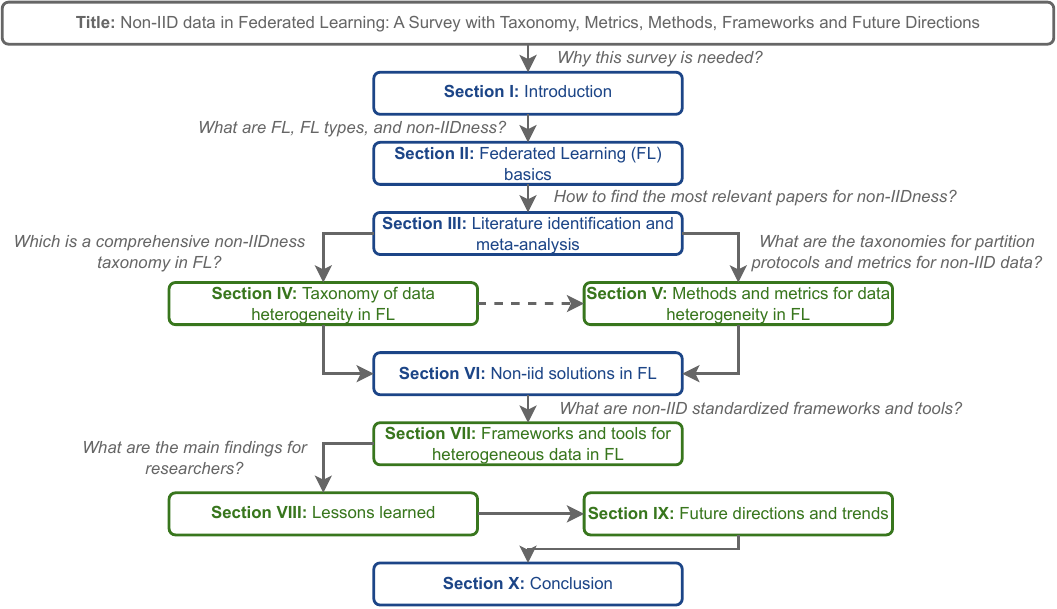}
  \caption{Outline of our survey}
  \label{fig:outline}
\end{figure*}

%% file: Sections/2.FL_basics.tex
\section{Federated learning basics}\label{sec:FL-basics}


FL is an innovative ML paradigm that enables training models on distributed datasets across multiple clients holding local data samples (a.k.a examples, individuals, data points) without exchanging them. This approach contrasts traditional centralized ML techniques, where all the data get aggregated in one location for model training.

FL minimizes a global objective function $l(w)$, represented as a weighted average across the private datasets of all participating clients. In this framework, we consider a scenario with $K$ total clients, where each client $k$ possesses its private dataset $\mathcal{D}_k$. Equation Eq.~\ref{eq:fl_loss} defines the function minimized,

\begin{equation} \label{eq:fl_loss}
\begin{aligned} 
     \min_{w} l(w) := h(L_k(w))
\end{aligned}
\end{equation}

where $w$ represents the parameters (a.k.a. weights) of the global model trained, and $h$ is function to aggregate the clients parameters. $L_k(w)$ is a local objective function such as cross-entropy loss for a supervised classification task~\cite{mora2024enhancing}.


The core idea behind FL is to train the global model $w$ by aggregating the local model updates (e.g., model gradients or weights) from the participating clients. The latter occurs through a process called Federated Averaging (\emph{FedAvg})~\cite{mcmahan2017communication,sun2022decentralized}, where each device trains its local model on its data, and the central server aggregates the model updates from the devices to update the global model. In FedAvg the aggregation function $h(L_k(w))=\sum_{k=1}^{K}{\frac{n_k}{n}L_k(w)}$, where $w$ represents the parameters of the global model trained, $n_k$ represents the size of $\mathcal{D}_k$, and $n$ is the total number of samples held by all participating clients. The mentioned iterative process of local training and global model aggregation continues for a pre-defined number of communication rounds ($T$), allowing the model to be trained on a diverse range of data sources while preserving the privacy of the individual data contributors, as the raw data never leaves the local devices. Alternative aggregation functions to FedAvg in FL are explained in Section~\ref{sec:non-iid-solutions}.

\subsection{Federated Learning types}

FL gets categorized into different types based on the participating entities. These participants can range from individual mobile devices and IoT sensors (cross-devices) to large organizations and institutions (cross-silos). The following paragraphs explain each type in more detail~\cite{mammen2021federated}.

\subsubsection{Cross-silos}
It involves collaboration between a limited number of organizations or institutions, typically with relatively powerful computational resources. In this setting, each client (or "silo") has a substantial amount of data and robust infrastructure. Examples include hospitals sharing medical data for research or banks collaborating on fraud detection models while maintaining client privacy~\cite{huang2022cross}. 

\subsubsection{Cross-devices}
Cross-device FL operates on a much larger scale, involving numerous edge devices such as smartphones, IoT devices, or wearables. This type of FL deals with many participants with limited computational resources and smaller, often intermittently available datasets. Cross-device FL must handle challenges like unreliable network connections, device heterogeneity, and frequent client unavailability. Its use is commonly in mobile keyboard prediction, voice assistants, and other consumer-facing applications~\cite{huang2022cross,nguyen2021federated}.

\subsubsection{Hierarchical FL}
Hierarchical FL introduces a multi-level structure to the FL process, typically involving cross-silo and cross-device elements. In this approach, devices or lower-level participants aggregate their models within local clusters or organizations before further aggregation occurs at higher levels. This structure can improve efficiency in large-scale systems, reduce communication overhead, and allow for more flexible data-sharing policies. Hierarchical FL is beneficial in scenarios involving multi-national corporations, healthcare networks, or large-scale IoT deployments~\cite{liu2020client}.

Regarding the data disposition, FL can be categorized primarily into horizontal and vertical partitions. The data disposition pattern influences aspects such as model architecture, aggregation methods, and privacy-preserving techniques employed in the FL system. The following paragraphs define each type thoroughly~\cite{zhang2021survey}.

\subsubsection{Horizontal FL}
Horizontal FL, also known as sample-based FL, is applicable when different participants have datasets with the same feature space but different sample sets. For instance, two regional banks might have the same types of customer data (features or attributes) but for different sets of customers (samples)~\cite{banabilah2022federated}. 

\subsubsection{Vertical FL}
Vertical FL, or attribute-based FL, is used when participants have datasets with identical samples but different feature spaces. This scenario often arises in multi-party business collaborations. For example, a bank and an e-commerce company might have different data types (attributes or features) for the same set of customers~\cite{liu2024vertical}. 

\subsubsection{Peer-to-peer FL}
Also known as decentralized FL, removes the central server, using peer-to-peer communication for model aggregation via epidemic or diffusion processes~\cite{roy2019braintorrent}. This approach improves scalability, fault tolerance, and privacy, making it ideal for edge networks or blockchain-based systems. Unlike hierarchical FL, it relies on flexible, pairwise, or group-based updates.


\subsection{Non-IID data in FL}

In centralized learning, non-IID data typically refers to variations within a single dataset that violate the assumption of IID~\cite{zhu2021federated,zhao2018federated}. The latter might include class imbalances, temporal shifts in data distribution, or biases in data collection. Although challenging, researchers often address such issues using data augmentation~\cite{maharana2022review}, resampling~\cite{avelino2024resampling}, or careful stratification during train-test splits~\cite{varoquaux2023evaluating}. In contrast, non-IID data in FL presents a more complex scenario. In FL, not only each participant's local dataset can be non-IID, but the data distribution across participants can also be highly heterogeneous. It leads to a two-level non-IID problem: intra-client non-IID (within each participant's dataset) and inter-client non-IID (across different participants). The decentralized nature of FL, combined with privacy constraints that limit direct data sharing, makes traditional centralized approaches to handling non-IID data insufficient~\cite{rahman2020internet}. 

\subsection{Downstream effects of non-IID data in FL}

Non-IID data in FL can significantly affect the learning process's performance, efficiency, and reliability. These consequences include but are not limited to:

\begin{itemize}
    \item \emph{Model bias:} The model may become biased towards the data distribution of participants with more prominent or influential datasets~\cite{zhang2024fedac}.
    \item \emph{Slower convergence:} Non-IID data can increase training time as the global model struggles to reconcile divergent local updates~\cite{wu2022node}.
    \item \emph{Reduced model accuracy:} The global model may underperform on specific subsets of underrepresented data in the federated dataset~\cite{li2022federated}.
    \item \emph{Instability in training:} Non-IID data can cause fluctuations in model performance across training rounds, making it difficult to achieve consistent improvements~\cite{lin2021semifed}.
    \item \emph{Challenges in participant selection:} Non-IID data can complicate choosing representative participants for each training round~\cite{rai2022client}.
    \item \emph{Communication inefficiency:} More communication rounds may be required to reach convergence, increasing network overhead~\cite{asad2021evaluating}.
    \item \emph{Privacy risks:} In some cases, non-IID data can make inferring information about individual participants' datasets easier through model updates~\cite{ma2020safeguarding}.
    \item \emph{Susceptibility to attacks:} Non-IID data can aggravate vulnerabilities to cybersecurity threats unique to FL, such as poisoning attacks, where adversarial participants introduce malicious updates to compromise the global model~\cite{xia2023poisoning}.

\end{itemize}

FL systems must employ tailored strategies like dynamic aggregation approaches, individualized modeling techniques, and resilient federated optimization methods to mitigate these issues~\cite{zhang2024addressing,tian2022wscc,pandey2022contribution}. These specialized solutions facilitate effective learning from diverse data distributions while preserving data privacy and maintaining computational efficiency. 

\begin{figure}[ht]
  \centering
  \includegraphics[width=\linewidth]{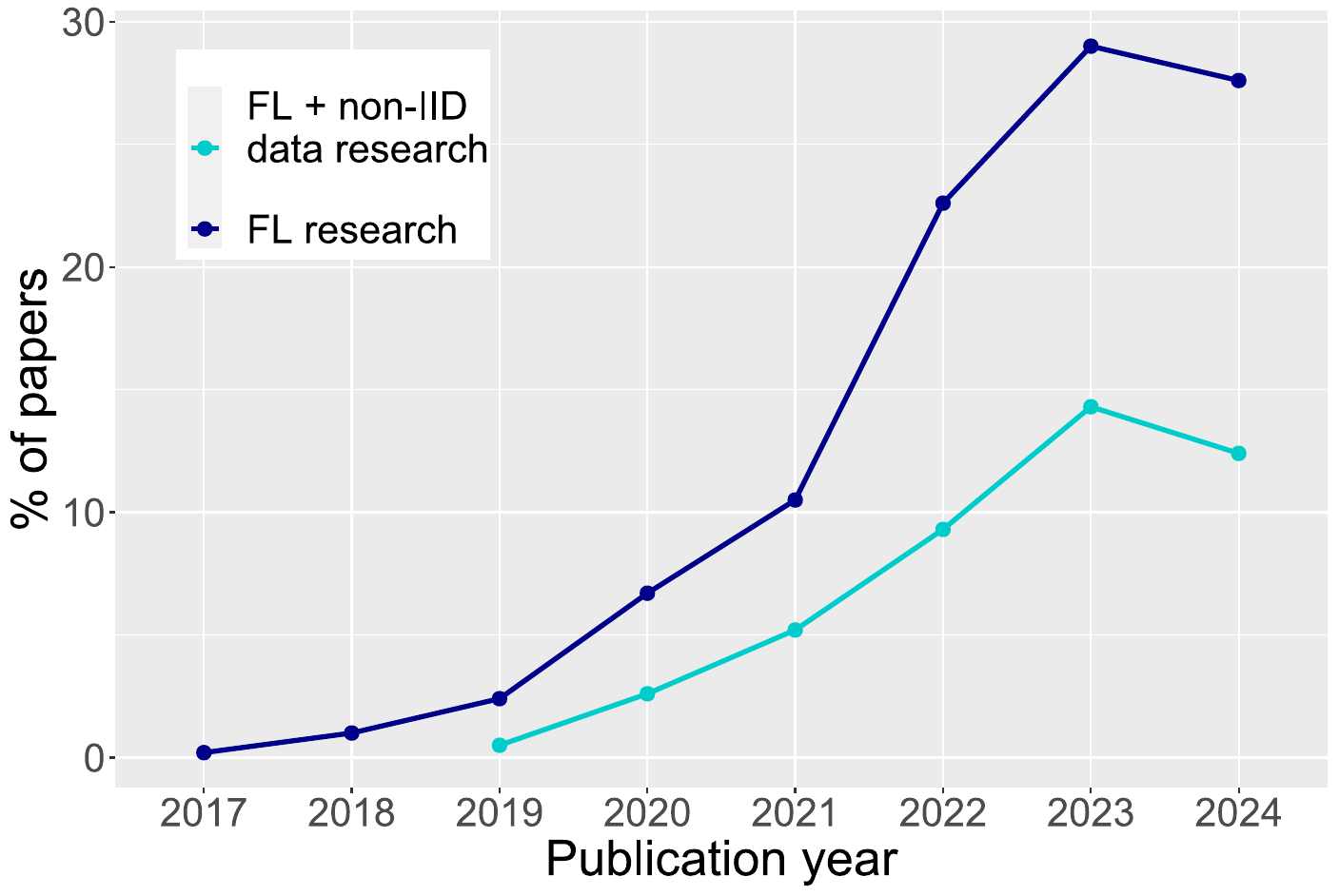}
  \caption{Research interest in general FL vs. non-IIDness in FL}
  \label{fig:fl_vs_noniid}
\end{figure}

The resulting challenges and complexities make non-IID data in FL relevant for researchers. In this context, Fig.\ref{fig:fl_vs_noniid} depicts how research focusing specifically on non-IID challenges in FL has notably increased since 2020, also peaking in 2023, highlighting the growing attention this issue is receiving. It is essential to emphasize that the apparent decrease in 2024 is due to the timing of our data collection, as papers were gathered during the third quarter, leading to incomplete data for that year (see Section~\ref{sec:SLR})). Thus, although general FL research continues to expand more quickly, the consistent rise in studies addressing non-IID data shows that it is emerging as a critical topic within the FL research community~\cite{silva2023towards}.

%% file: Sections/3.Literature_review.tex
\section{Literature identification and meta-analysis}\label{sec:SLR}
In this section, we outline the methodology used to identify and select the relevant papers and articles that form the basis of our survey. Additionally, we present a detailed meta-analysis of these papers, highlighting compelling information and insights.  

\subsection{Methodology for identifying relevant literature}

\begin{figure}[ht]
  \centering
  \includegraphics[width=\linewidth]{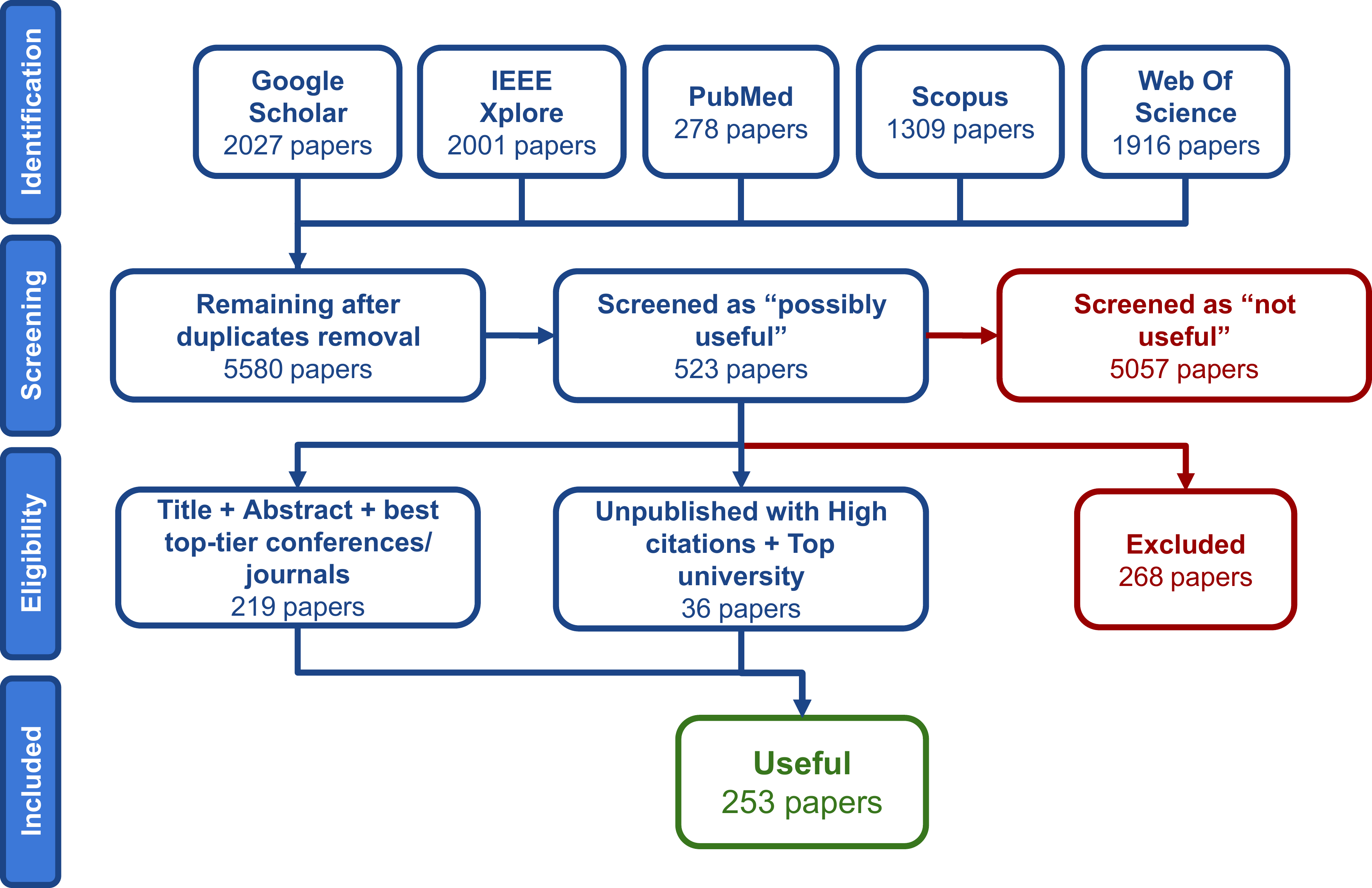}
  \caption{PRISMA flow for gathering relevant references}
  \label{fig:prisma_flow}
\end{figure}

We employed a methodology for identifying relevant literature based on the PRISMA methodology~\cite{siddaway2019systematic} to conduct a thorough and structured analysis of the current literature on non-IID-ness in FL. We improved the latter methodology by considering top-tier published papers and research from high-level universities. The approach involved gathering, filtering, and assessing relevant academic papers and resources from various scholarly databases. The process began with an \emph{identification} phase, as illustrated in Fig.~\ref{fig:prisma_flow}. We utilized six reputable academic repositories: Google Scholar~\cite{google-scholar}, IEEE Xplore~\cite{ieee-xplore}, PubMed~\cite{pubmed}, Scopus~\cite{scopus}, and Web Of Science~\cite{web-of-science}. We chose these platforms for their comprehensive coverage across multiple academic fields and their extensive collection of research publications. Our search strategy involved 83 queries across ten categories related to non-IID-ness in FL, including general non-IID-ness, label skew, attributes skew, quantity skew, distribution shift, non-IID metrics, performance metrics, partitioning protocols, datasets, and frameworks. We input each of the 83 queries into each academic repository to retrieve relevant papers. After eliminating duplicate documents, we compiled a collection of 5489 unique papers for further analysis.

We screened each article after the initial collection to identify potentially valuable papers. Our \emph{screening} process consisted of the following steps for each paper retrieved:

\begin{enumerate}
    \item We examined the title and available keywords of every paper.
    \item If these elements were relevant (for instance, if they addressed non-IID-ness, discussed heterogeneity, mentioned label skew, or compared centralized and federated approaches), we marked the paper as "possibly useful" for further review.
\end{enumerate}

The latter allowed us to filter the large pool of papers, focusing on those most likely to contribute to our study of non-IID-ness in FL.

In the \emph{eligibility} phase, we refined the selection of references to focus on the most relevant and impactful works related to non-IID-ness in FL by reviewing the titles and abstracts and identifying the conference ranking or journal quartile. As a result, 202 papers were marked as "useful" if they met either of the following criteria:

\begin{enumerate}
    \item The title and abstract address non-IID-ness.
    \item The paper was published at a conference ranked A* or A or in a Q1 journal.
\end{enumerate}

Some relevant papers are often not published in conferences or journals. Still, it is worth analyzing whether they have received many citations or come from high-ranking universities. Then, 36 additional papers are marked as "useful" if it was not published in a journal or conference (i.e., published in Arxiv) and any of the following occurs:

\begin{enumerate}
    \item For papers from 2024, it has been cited (using the normalized-by-year citations count) at least once. 
    \item For papers before 2023 included, it was cited more than four times for other years (using the normalized-by-year citations count).
    \item The author's universities included in the paper are in the top 100 universities regarding the QS ranking - Engineering and Technology|\cite{qs-ranking}.
\end{enumerate}

After applying the rigorous eligibility process described, we \emph{included 235 papers} to serve as the foundation for our survey.

\subsection{Meta-analysis of relevant literature}

We comprehensively examined the 235 papers selected through our process to provide a general glance at the current research on non-IID-ness in FL. Through this analysis, we seek to contextualize individual findings within the broader landscape of non-IID FL literature and provide a foundation for understanding the collective progress and challenges in this rapidly evolving domain.



\begin{figure}[ht]
  \centering
  \includegraphics[width=\linewidth]{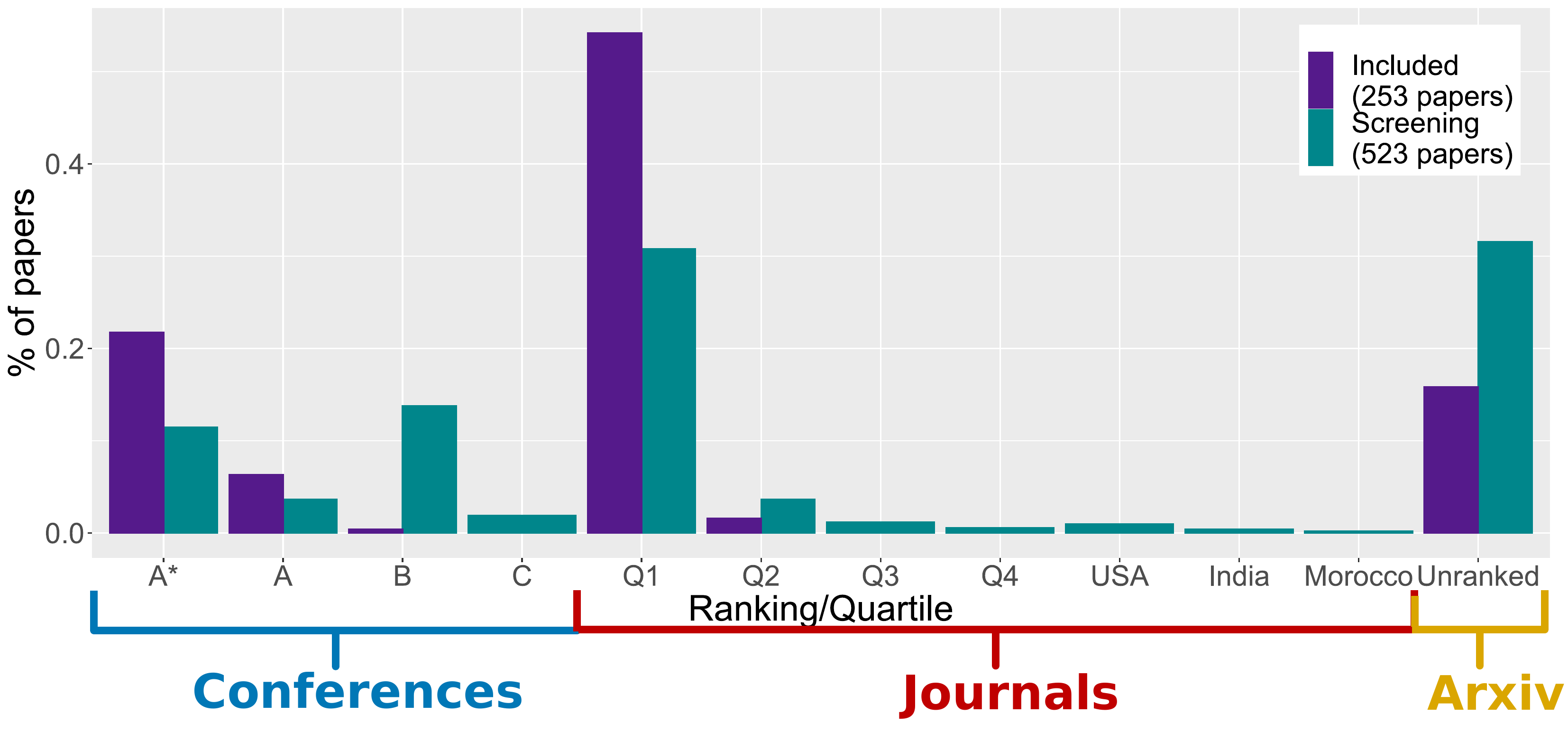}
  \caption{Rankings and quartiles distribution comparison}
  \label{fig:ranking_quartile_compare}
\end{figure}

Fig.~\ref{fig:ranking_quartile_compare} presents a comparative analysis of the publication quality between the papers included in our final study and those initially retrieved during the screening phase. It demonstrates a clear shift towards higher-quality publications in our selected ones. Specifically, the included papers predominantly originate from top-tier conferences (A* and A rankings) and high-impact journals (Q1 quartile). This notable difference in quality distribution underscores the efficacy of our rigorous selection criteria and eligibility process. 

\begin{figure}[ht]
  \centering
  \includegraphics[width=\linewidth, trim={2cm 3cm 2cm 12cm},clip]{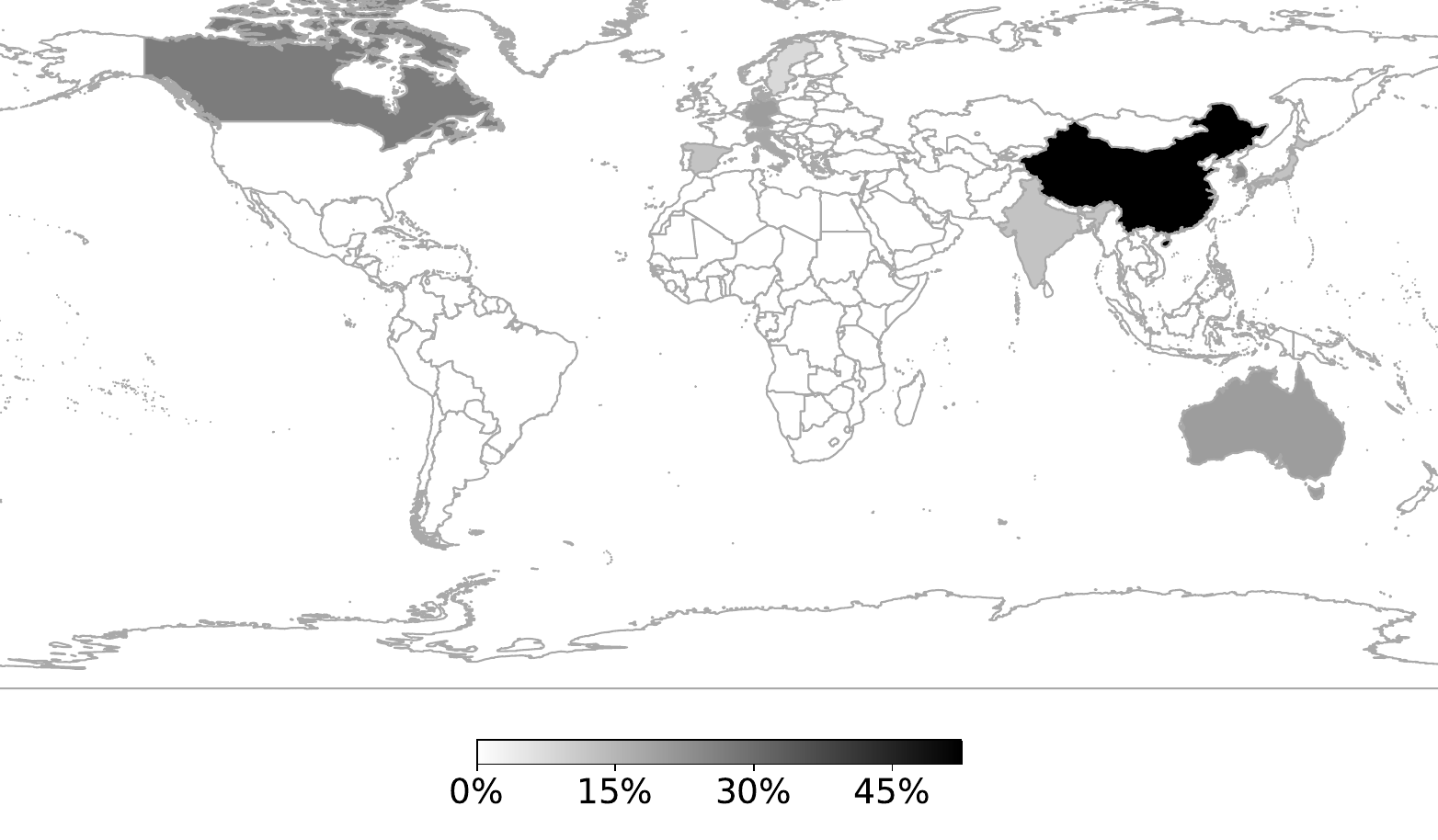}
  \caption{Geographic distribution of collected papers by first author affiliation country. Map values indicate each country's relative prevalence inside the collected database.}
  \label{fig:paper_choropleth}
\end{figure}

The world map of Fig.~\ref{fig:paper_choropleth} depicts the geographic distribution of collected papers based on the first author's affiliation. Highlighted in black, China emerges as the dominant contributor, indicating its leading role in this research field. North America (the United States and Canada) and Australia are the second most significant contributors. Other Asian countries (i.e., South Korea, Singapore, and Japan, among others) and Europe (Germany, Italy, Spain, etc.) are represented in lighter shades with less research participation. Those countries in white represent the lack of research contribution. This map reveals a precise concentration of research output in specific regions, potentially reflecting global differences in research focus, funding, or technological infrastructure. 

\begin{figure}[ht]
  \centering
  \includegraphics[width=0.6\linewidth]{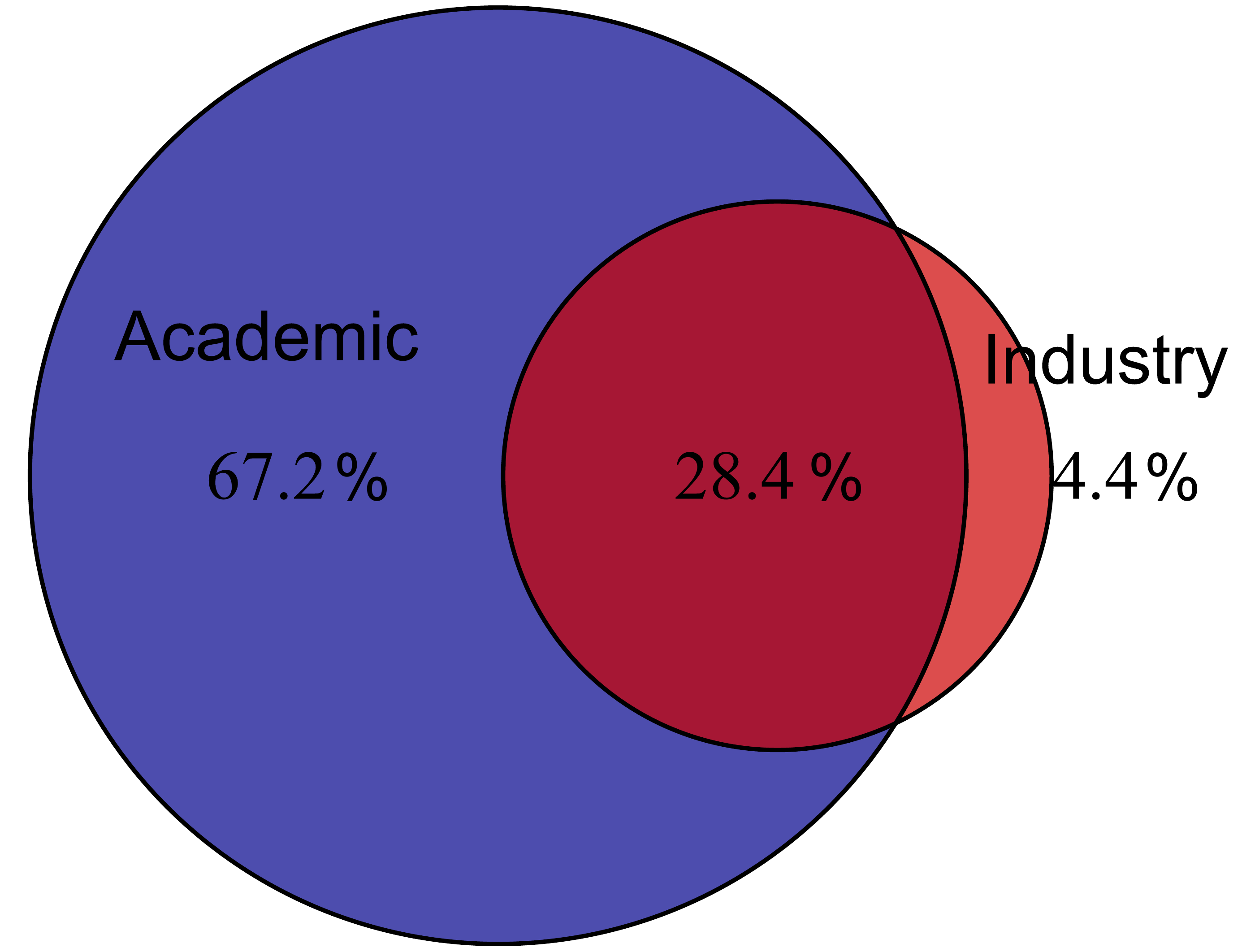}
  \caption{Percentage of research done in academic and industrial environments}
  \label{fig:academy_industry}
\end{figure}

Fig.~\ref{fig:academy_industry} illustrates the distribution of research conducted in academic versus industrial environments. It reveals that academia dominates the research landscape, accounting for 65.1\% of the studies. Interestingly, there's an overlap of 30.6\% between academic and industrial research, suggesting a considerable amount of collaborative work or researchers with dual affiliations. A small segment of 4.3\% represents research conducted solely in industry. It underscores the importance of academic institutions in driving research in non-IIDness while highlighting the industry's notable role, mainly through collaborations with universities.

%% file: Sections/4.Taxonomy_non_iid.tex

\section{Taxonomy of data heterogeneity in FL}
\label{sec:data-heterogeneity}



\begin{figure*}[ht] 
    \centering
    \includegraphics[width=\textwidth]{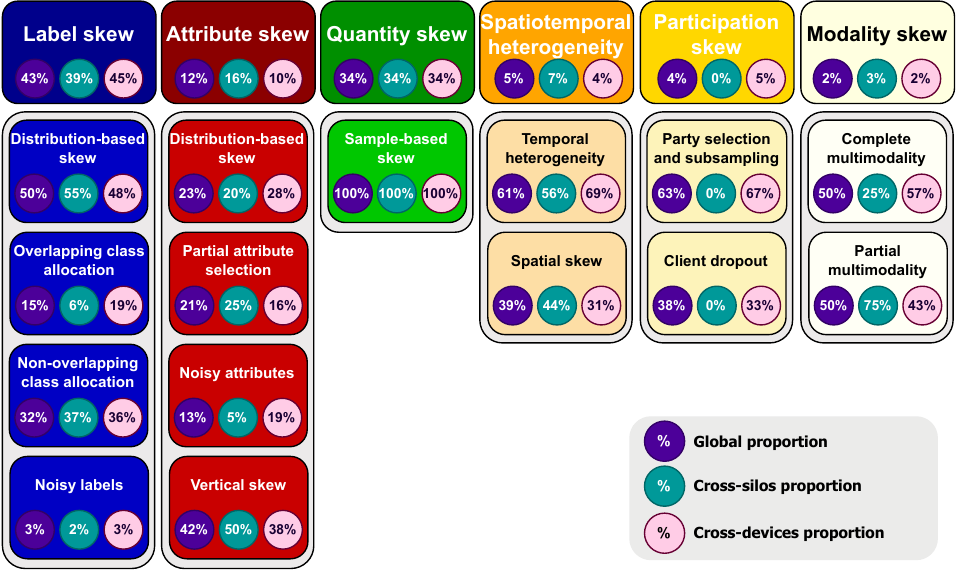}
    \caption{Taxonomy for non-IIDness in FL. The figure categorizes different types of data skews and heterogeneities. Percentages indicate the proportion of each type across global, cross-silo-related, and cross-device-related papers.
    }
    \label{fig:taxonomy}
\end{figure*}

To define different data skews, we assume in this section that all the relevant densities exist for simplicity. We can generally replace densities, e.g., with cumulative density functions below, as long as all the relevant conditional distributions are well-defined. 
For general terminology, we use \emph{skew} to refer to various differences between clients' local data sets, and \emph{heterogeneity} as a more general category that might or might not imply differences between clients.


Writing $\mathcal D_i \sim \mathbb D_i$  for the $i$th clients' local data, where $\mathbb D_i$ represents the underlying data distribution from which the $i$th client's local data $\mathcal D_i$ is sampled, and denoting the probability density  function (pdf) of $\mathcal D_i$ by $f^{(i)}$, \textbf{data skew} means that 
\begin{equation}
    f^{(i)} \not= f^{(j)} \text{ for some i,j } \in \{1,\dots,K\} .
\end{equation}
In other words, this means that the local data distributions differ somehow between at least one pair of clients. Data skew is a challenge in FL because it can lead, e.g., to drift or bias in the local models, complicating the aggregation process and hindering the performance and generalization of the global model across clients.
\footnote{Note that data skew is a theoretical concept that pertains to the local data-generating processes. It is not directly observable with finite empirical datasets, and testing for it involves inherent uncertainty. On a practical level, existing research uses various methods to simulate data skew empirically, which we also cover in the following.}



In the supervised learning setting with $\mathcal D_i = (y_i, x_i)$, where $y_i$ are the labels (target outputs or classes assigned to samples) and $x_i$ the attributes (features or properties describing samples), we can further divide general data skew into label and attribute skews as in the following:
\begin{enumerate}
    \item {
        \textbf{Label skew.} 
        We have \emph{marginal} or \emph{conditional label skew}, respectively, if 
        \begin{align}
            \FY{i} \not =& \FY{j} \text{ or } \\
            \FYX{i} \not =& \FYX{j} 
            \text{ for some clients i,j}
        \end{align}
        i.e., the marginal $\FX{i}$ or conditional $\FYX{i}$ distributions of the labels differ between some pairs of clients. In general, with heterogeneous data we might find either marginal or conditional skew or both. However, if there is no attribute skew (see below), then marginal label skew implies conditional label skew and vice versa. 
        %
        %
        More formally, 
        assuming $\FX{i} = \FX{j}$ and $\FXY{i} = \FXY{j}$ (no attribute skew) implies 
        $\FY{i} \not = \FY{j} \Leftrightarrow \FYX{i} \not = \FYX{j}$.
            The proof follows immediately from Bayes' Theorem:
            \begin{equation}
                \FYX{i} = \frac{ \FXY{i} \FY{i} }{\FX{i} } 
            \end{equation}
        which readily implies the claim when comparing to client $j$ and canceling out the common factors.

        As illustrated by Fig.~\ref{fig:taxonomy}, we further divide label skew 
        into four different subtypes based on the analyzed papers. 
        \emph{(1) Distribution-based skew} refers to a situation where clients share classes but the prevalence of each class is ruled by some defined distribution (e.g., client A might have 80\% cats and 20\% dogs, while client B has 30\% cats and 70\% dogs)~\cite{wu2024fedel,liu2024aedfl,li2024federated,qiao2024towards,li2023effectiveness,dai2023tackling,yan2023label,zhang2022federated,he2022learning,zhang2021fedpd}. \emph{(2) Overlapping class allocation} describes scenarios where multiple clients share some common classes but each client may also have some unique classes (e.g., client A has classes 1,2,3 while client B has classes 2,3,4)~\cite{yao2024ferrari,cheriguene2023towards,jamali2022federated,huang2021personalized}. 
        \emph{(3) Non-overlapping class allocation} represents a more extreme case where different clients have completely disjoint sets of classes with no overlap between them (e.g., Client A has classes 1,2 while Client B has classes 3,4)~\cite{diao2024exploiting,ma2023flgan,shen2023ringsfl,peng2022byzantine}. 
        Finally, \emph{(4) Noisy labels} refers to cases where some data points in clients' datasets are incorrectly labeled, adding noise to the FL process~\cite{jebreel2024lfighter,zhang2024cross,jin2023blockchain}.
        
    }
    \item{
        \textbf{Attribute skew.} 
         Similarly as with the labels, we can look at the marginal and conditional distributions of the attributes and have \emph{marginal attribute skew} or \emph{conditional attribute skew}, respectively, when 
        \begin{align}
            \FX{i} \not =& \FX{j} \text{ or } \\
            \FXY{i} \not =& \FXY{j} 
            \text{ for some clients i,j} .
        \end{align}
        Mirroring the situation with the labels above, if there is no label skew, then marginal attribute skew implies conditional attribute skew and vice versa. 

        As depicted in Fig.~\ref{fig:taxonomy}, we divide attribute skew 
        into four subtypes based on our analysis of the existing research. 
        \emph{(1) Distribution-based skew} refers to situations where clients share the same features but have different probability distributions over these features (e.g., different distributions of pixel values in images across clients)~\cite{li2024fedcir,wang2024one,ezzeldin2023fairfed,huang2022fairness,onoszko2021decentralized,reisizadeh2020robust}. 
        \emph{(2) Partial attribute selection} describes scenarios where clients have incomplete or partially observed features for their samples~\cite{badar2024fairtrade,chen2023fraug}. 
        \emph{Noisy attributes} represents cases where the feature values in some clients' datasets contain noise or corrupted values, affecting the data quality~\cite{he2023clustered,nguyen2023federated}. 
        \emph{(4) Vertical skew}, also known as vertical FL, refers to situations where different clients have different subsets of features for the same samples (e.g., Client A has features 1-10 while Client B has features 11-20 for the same set of instances)~\cite{tan2023federated,yan2023simple,chiaro2023fl,das2022cross,jones2022federated}.

        }
    \end{enumerate}


\hlmycolor{Unlike most previous surveys, we include the notion of \textbf{Modality skew} under the taxonomy of non-IIDness}. It is a specific type of data skew referring to a setting where different clients have data from varying input modalities, such as text, images, audio, tabular, or sensor data, as opposed to all clients having data of the same type. This occurs when clients collect or generate data in distinct formats, leading to heterogeneity in the input data types across the network. Modality skew is a subtype of the attribute and label skews whose interdependence is detailed in Fig.~\ref{fig:interdependencies}.

As represented in Fig.~\ref{fig:taxonomy}, we distinguish between two subtypes of modality skew based on our analysis. 
\emph{(1) Complete multimodality skew} refers to a scenario where all clients have the same types of multimodal data (e.g., all clients have both image and text data), but there might be differences in the distribution of these modalities across clients~\cite{feng2023fedmultimodal,chen2022towards}. For example, one client might have a different distribution of image-text pairs compared to another client while still maintaining all modalities. 
In contrast, \emph{(2) partial multimodality skew} represents a situation where clients have varying subsets of the possible modalities (e.g., some clients might have both image and text data, while others might only have image data)~\cite{borazjani2024multi,ouyang2023harmony}.


Besides the main forms of data skew discussed above, which are based on differences in the local data distributions between clients, we further identified quantity skew, spatiotemporal heterogeneity, and participation skew as contributing to the general data heterogeneity in FL. However, as detailed next, these forms of heterogeneity differ in essential ways from the forms of data skew defined above.

\begin{enumerate}
    \item \textbf{Quantity skew.} Another important factor of data heterogeneity in FL is the possibly wildly different amount of data each client holds. We call this type of data heterogeneity \emph{quantity skew}. Unlike data skew, the definition of which is based on the local data generating processes, 
    quantity skew is a purely empirical quantity. It is orthogonal to data skew, and the actual effect depends on whether there is data skew present or not.

    As shown in Fig..~\ref{fig:taxonomy}, quantity skew only contains the subtype called \emph{sample-based skew}, which refers to differences in the number of samples among the clients~\cite{zhou2024fault,cong2024ada,casella2024experimenting,mu2023fedproc,wang2023distribution,morafah2023flis,lian2022blockchain,de2022mitigating,xiong2021privacy}. Moreover, our analysis of the literature identified a related concept known as \emph{data sparsity}. This term describes a specific form of sample-based quantity skew, where certain clients systematically lack data for specific features, classes, or patterns~\cite{mao2023safari,liao2023adaptive,qiu2022zerofl,huang2022achieving,tong2020federated}. For instance, in the context of medical records, data sparsity might manifest as some clients frequently having missing blood test values or incomplete patient histories. As such, data sparsity can be interpreted as a specific realization of data skew.

        \item \textbf{Spatiotemporal heterogeneity.}
        This type of heterogeneity arises when data across clients differs in either or both spatial or temporal dimensions. 
        As presented in Fig.~\ref{fig:taxonomy}, following here the common trend in the analyzed papers, by \emph{(1) temporal heterogeneity} we simply mean that the local data on a given client form a time series, i.e., the local samples on a given client have a temporal dimension instead of being IID samples~\cite{li2024facing,tan2024heterogeneity,shen2024decentralized}. 
        This is orthogonal to having data skew in the above sense, as the stochastic process generating the local data might be the same (implying no data skew) or different between clients. 
        When the learning algorithm or the clients only have access to data corresponding to a single time-step at a time, this can lead to the federated continual learning problem~\cite{yang2024federated,casado2023ensemble}.

        In turn, and again following the usage in the analyzed papers, by \emph{(2) spatial skew} we refer to differences in client data distributions (i.e., having some form of data skew) caused by differences in the physical location of the clients~\cite{al2024fedagat,huang2023train,chen2023spatial}. 

        \item \textbf{Participation skew.} We refer to imbalances in client participation during the training or testing/production phases in FL as \emph{participation skew}. 
        This skew arises from two key factors: client selection, i.e., from how and when the clients are included in the training, and client dropouts, where some otherwise included clients may intermittently fail to participate in training due to network issues, device unavailability, or other constraints.
        
        As a result of participation skew, not all clients contribute equally or consistently to model training. In the presence of data skew, such imbalances can give rise to biases in the global model that reflect the over-representation of certain clients while others are underutilized or excluded. This can lead to various issues, such as different forms of biases (for example, when some group of clients has systematically worse utility) and problems with out-of-distribution (OOD) generalization (e.g., when there is data skew between training and testing/production set of clients).

        Regarding Fig.~\ref{fig:taxonomy}, participation skew occurs in two types that we define according to the analysis of the papers. 
        \emph{(1) Party selection and subsampling} refers to the strategic or random selection of a subset of clients to participate in each training round, which can be influenced by factors such as computational resources or communication constraints (e.g., only selecting 10 out of 100 available clients per round)~\cite{jin2024performative,yang2021client}. 
        \emph{(2) Client dropout} describes scenarios where clients unexpectedly become unavailable or disconnect during the training process, which can happen due to network issues, device limitations, or other technical problems (e.g., a client might lose connection mid-training or run out of battery)~\cite{jeon2024federated,wang2024fluid}. 
        
\end{enumerate}

\subsection{Prevalence of non-IIDness types in FL}


As shown in Fig.~\ref{fig:taxonomy}, \hlmycolor{label skew is the most prevalent type of heterogeneity studied in the existing literature} with a higher prevalence in cross-device than in cross-silo settings. 
Quantity skew is the second most studied type, with the same prevalence in cross-silo and cross-device studies. 
%
Despite its importance, \hlmycolor{spatiotemporal and especially temporal heterogeneity is not broadly treated in the papers analyzed}. 
We additionally note that modality skew is an understudied topic, with only 2\% of the papers including it.

Looking at the subtypes within each category reveals interesting patterns. For instance, within label skew, distribution-based skew (48-55\%) and non-overlapping class allocation (36-52\%) are much more common than overlapping class allocation (9-19\%) and noisy labels (2-3\%). In the attribute skew category, vertical skew (38-50\%) dominates other subtypes. The latter occurs because vertical skew relates directly to vertical FL (see Section~\ref{sec:FL-basics}) which is relatively more prevalent in real-world datasets than the other subtypes. 
%
%
For participation skew, party selection and subsampling as well as client dropout show significant variation across contexts (0-67\%, 0-38\%, respectively). This is mainly explained by the fact that neither form of participation skew is typically relevant for cross-silo studies, as in most cases all clients are included on every training and testing round, and the clients are assumed to have good connectivity. 

These differences in prevalence suggest that certain types of non-IIDness have been studied much more than others. For example, label skew is significantly more common than attribute skew in the papers we have analyzed. This could guide researchers and practitioners in prioritizing which challenges to address in research as well as in designing practical FL systems. 
%


\subsection{Interdependence of skew and heterogeneity types}

In previous sections of this manuscript, we provide an \emph{individual overview} of the various types of data heterogeneity that can occur in FL. \hlmycolor{In practice, the various types of data heterogeneity in FL do not operate in isolation. Instead, they can exhibit complex interdependencies}, where one form of heterogeneity can directly cause or modify the impact of another.
%

\begin{figure}[ht] 
    \centering
    \includegraphics[width=\columnwidth]{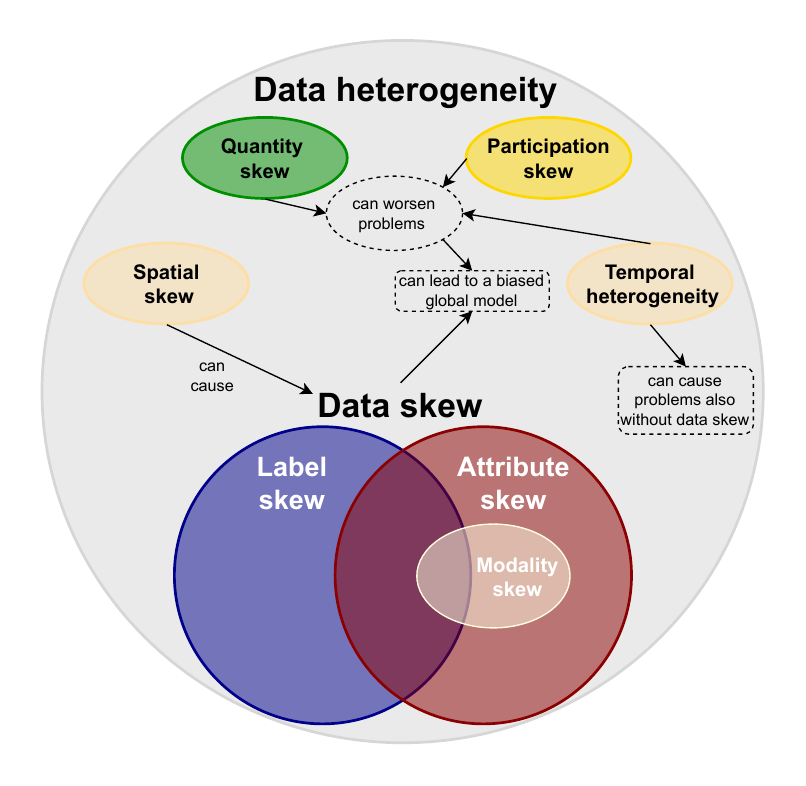}
    \caption{Interdependencies between skew types in FL 
    \label{fig:interdependencies}}
\end{figure}

As illustrated in Fig.\ref{fig:interdependencies}, data skew encompasses both label skew and attribute skew, which can occur independently or simultaneously, such as bank A having mostly legitimate transactions (label skew) and bank B having high-value transfers (attribute skew). Additionally, modality skew represents a subset of attribute skew and may coincide with label skew, as occurs when Hospital A has X-rays of pneumonia cases while Hospital B has computed tomography scans of healthy patients.

On the other hand, other types of heterogeneity may influence, induce, or worsen the problems generated by these core skew types. As an example, running FL training with clients physically located in different places (spatial heterogeneity) can cause data skew between clients, which can further combine with participation skew (e.g., when only recharging mobile devices can take part in training and charging is more common during clients' night time) to induce notable biases in the global model. Another related concept is temporal heterogeneity, which can make learning harder also in the FL setting even without any data skew.

Furthermore, any form of data heterogeneity possible in the centralized data settings can also appear in the FL context. As an example, there might be differences in the distributions between training and testing/production data sets, which can then lead to OOD generalization issues with centralized data. 
Similarly, differences between training and test/production data sets might appear in FL, either systematically inside each client's data, in which case there need not be any between-clients data skew or with data skew. 
Such combinations can arise in practice, e.g., due to the temporal difference between training a model and using it in production, even when the sets of training and production clients would be essentially the same.

Tackling these interrelated skews demands a comprehensive approach that accounts for how each type contributes to data heterogeneity and affects the learning performance in federated models.

%% file: Sections/5.Protocols_and_metrics.tex
\section{Methods and metrics for data heterogeneity in FL}\label{sec:protocols-and-metrics}


This section provides state-of-the-art standardized partition protocols to split centralized data into federated datasets and metrics to quantify non-IIDness among clients.

\subsection{Standardized dataset's partition protocols}

Protocols for dataset partitioning in FL are crucial for advancing research and practical applications. \hlmycolor{The current lack of \emph{consensus} on the sufficient conditions and scenarios for testing solutions to non-IID data has led to an increase of partitioning protocols and experimental setups}. This inconsistency complicates the replication of results, validation of findings, and comparison of different approaches. Establishing agreed-upon methods for creating, quantifying, and evaluating non-IID partitions is essential for enabling fair comparisons, enhancing reproducibility, and providing a common framework for describing various non-IID scenarios. 

\begin{figure}[ht] 
    \centering
    \includegraphics[width=\columnwidth]{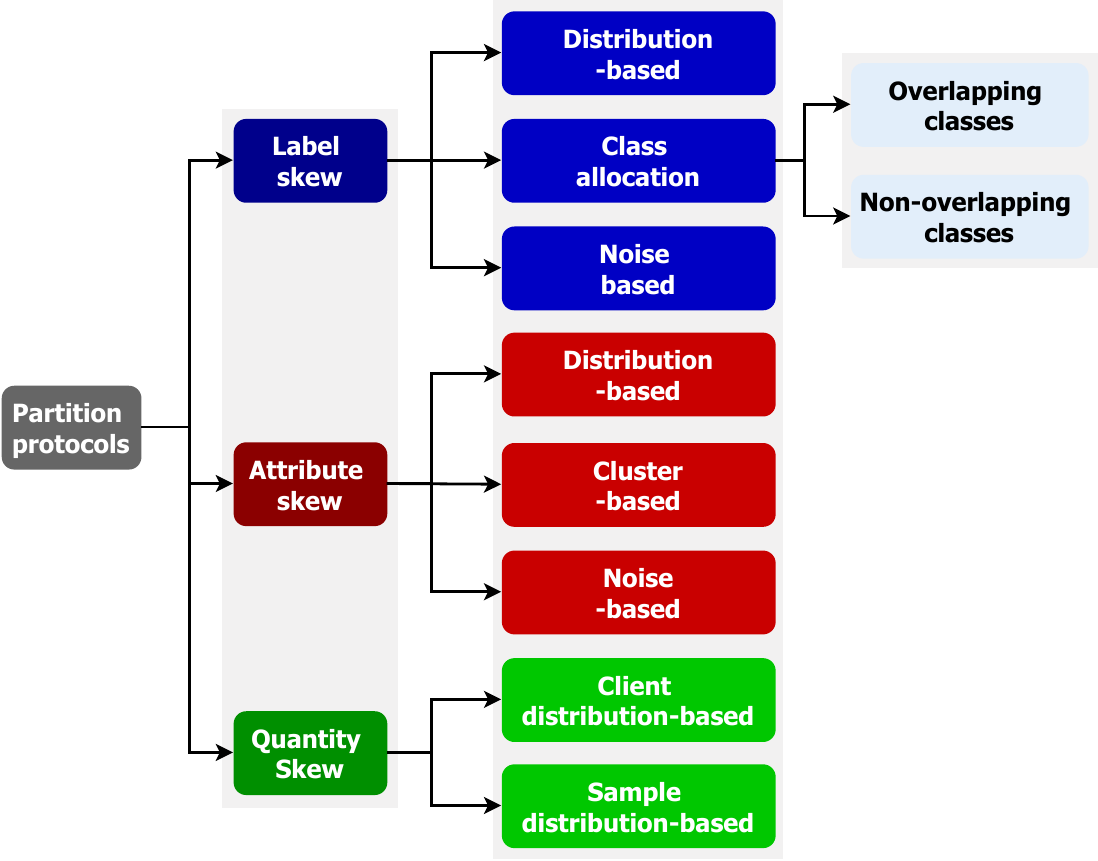}
    \caption{Taxonomy for partition protocols in Federated Learning}
    \label{fig:taxonomy_protocols}
\end{figure}

Leveraged on the exhaustive analysis of the selected papers from the survey, we created a taxonomy for the partition protocols included in the recent literature. Fig.~\ref{fig:taxonomy_protocols} depicts the classification encountered for the methods used in the non-IID research field.

In the following paragraphs, we define each category of the taxonomy presented in Fig.~\ref{fig:taxonomy_protocols}, together with the explanation of the detailed methods to partition centralized data into federated data.

\subsubsection{\textbf{Label skew}} It refers to partitioning the centralized dataset considering the uneven distribution of class labels across different clients in an FL setting as exposed in Section~\ref{sec:data-heterogeneity}. It is divided into three subcategories: Distribution-based, Class allocation, and Noise-based.

\paragraph{\emph{Distribution-based}} This approach allocates data to clients based on a specified probability distribution of labels to create imbalanced label distributions across clients.

The \emph{Dirichlet} partition protocol is the most employed in research, with 27\% of the papers implementing it~\cite{song2024feddistill,wang2024fedbnr,li2024aligning,lian2023gofl,zheng2023poe,you2022reschedule,huang2022fairness,yao2021local,guo2021towards}. It uses the Dirichlet distribution (DD)~\cite{minka2000estimating} to allocate samples across clients, simulating non-IID scenarios. A concentration parameter $\alpha$ controls the degree of data heterogeneity. Lower values create more skewed distributions, while higher values approach IID conditions. This method permits generating a range of realistic non-IID datasets for evaluating FL algorithms under varying degrees of data imbalance. The DD generalizes the Beta distribution, which generalizes the Uniform distribution. As a result, it leads to a skewed data split~\cite{lin2016dirichlet}.

Another method in this category is the ~\emph{Generator of Non-IID datasets}~\cite{zhou2022towards}. It receives as inputs 
the minimum and maximum number of data points that can be assigned to any partition ($D_{\text{min}}$, $D_{\text{max}}$), the number of labels (or classes) to be sampled for each partition ($L_{\text{num}}$) and the number of partitions ($P$). Then, it creates a set of data partitions $\text{Dataset} = (\mathcal{D}_1, \mathcal{D}_2, \dots, \mathcal{D}_P)$. For each partition $i$, it randomly samples $L_{\text{num}}$ labels from the set of unique categories. It then selects a random number $D_{\text{num}}$ within the range $[D_{\text{min}}, D_{\text{max}}]$. A set of weights is sampled from the interval $(0,1)$, normalized by dividing each by the sum of all weights. Each class's total number of data points is calculated by multiplying $D_{\text{num}}$ by these normalized weights. Finally, data is sampled according to this computed number of data points and the chosen classes, generating the non-IID datasets for each partition.

\paragraph{\emph{Class allocation}} This method assigns entire classes or datasets of classes to different clients. It has two variations:

\subparagraphnumbered{\emph{Overlapping classes:}} In this variation, some classes are shared among multiple clients, while others may be unique to specific clients.

The \emph{Percentage-of-non-IID-ness} method~\cite{jimenez2024fedartml,sun2024understanding,arafeh2023data,noble2022differentially,hsieh2020non} is the one employed the most in this category, reaching 7\% of the papers. It controls the skewness by adjusting the fraction of data that is non-IID. For example, 20\% non-IID indicates that 20\% of the dataset is partitioned based on labels, while the remaining 80\% is partitioned uniformly at random. While varying the percentage, different levels of skewness in the dataset distribution can be achieved, allowing for controlled experiments with different degrees of heterogeneity.

Another method is the \emph{Dominance class Ratio}~\cite{zhang2024addressing,shi2023fairness,ye2023fedfm,zhao2021federated,chen2021fedsa}, which has a statistical parameter $\sigma \in [0, 1]$ to control the ratio of the dominant class within each client. The latter indicates that one class primarily dominates within each client, while the remaining classes are uniformly distributed across the clients. Adjusting $\sigma$ permits simulating different degrees of class dominance and distribution imbalances.

 One alternative method falling in this category is the ~\emph{local long-tail (LLT)} partition, which simulates Non-IID datasets with a long-tail distribution ~\cite{tang2021data,wang2020optimizing}. This method divides a global dataset $\mathcal{D}$ with $C$ classes among $K$ clients (where $K = C$). The parameter $\alpha_l$ controls the degree of non-IID-ness. For each class $c \in C$, the samples get partitioned into $K$ parts, with one part containing $\alpha_l N_{\ast,c}$ samples assigned to a specific client $k$ and the remaining $K-1$ parts containing $\frac{1 - \alpha_l}{K-1} N_{\ast,y}$ samples distributed to other clients. The latter ensures each client has one dominant class and $C-1$ tail classes. The LLT partition reflects real-world scenarios, like smartphone users favoring certain photo styles, but may not fully capture the distribution characteristics seen in real-world FL tasks. Despite this, the LLT method mirrors imbalanced learning scenarios, serving as a basis for testing data resampling effects in FL.

\subparagraphnumbered{\emph{Non-overlapping classes:}} Regarding this classification, each client is assigned distinct classes, with no overlap between clients.

The most used partition protocol in this category is the \emph{Sharding} method~\cite{zakerinia2024communication,wang2024feddbo,yu2023federated,zhang2023robust,zhang2021federated,singh2020fair}, employed by 20\% of the papers. It is a technique used to create non-IID datasets by sorting them based on class labels and then dividing them into smaller datasets (a.k.a shards) of equal size. These shards are distributed among clients, each receiving a fixed number. Since each shard contains data from only a few classes, clients have data representing only a small subset of the overall class distribution. This method results in a highly skewed, non-IID distribution, as clients predominantly possess samples from just a few classes, and these do not overlap among the clients. Notice that the case when each client holds only one class is known as~\emph{Pathological partition}~\cite{wang2024age,wang2024feddbo,shu2022clustered}.

The ~\emph{Archetypes} method~\cite{zhu2022decoupled,zhang2021client} divides the dataset into distinct groups, or archetypes, where each group contains data corresponding to a subset of class labels. Clients are assigned to one or more archetypes, meaning that each client only has access to a specific subset of the overall class distribution. For instance, one archetype may consist of data with labels \{0, 1, 2, 3, 4\}, while another includes labels \{5, 6, 7, 8, 9\}. This method results in a structured Non-IID distribution where clients' data is restricted to a limited range of classes.

\paragraph{\emph{Noise-based}} This technique introduces label noise to simulate label skew, including Symmetric and Asymmetric noise.

The ~\emph{symmetric}~\cite{jin2023blockchain,yin2023defending,gupta2022fl}, and ~\emph{asymmetric}~\cite{zhang2024cross} label noise methods introduce noise into the labels to simulate mislabeling. In the symmetric case method, the correct labels of each class are randomly flipped to any of the other categories with a fixed probability, ensuring an even distribution of incorrect labels across all other classes. In contrast, the asymmetric method involves mapping the correct labels to a specific confusion class, which is more likely to be mistaken for the original label. This results in a more structured type of noise, where mislabeling follows a particular pattern based on class similarities.

\subsubsection{\textbf{Attribute skew}} It refers to the simulation of uneven distribution of features or attributes across clients starting from the centralized data. It is divided into three subcategories: Distribution-based, Cluster-based, and Noise-based.

\paragraph{\emph{Distribution-based}} This approach creates variations in attribute distributions across clients using different parameters for defined probability distributions.

One method falling into this category is the ~\emph{Gaussian distribution affine} method~\cite{reisizadeh2020robust}, which introduces a distribution shift to the training samples at each client by applying an affine transformation. This transformation is randomly generated based on a Gaussian distribution, resulting in variations in the data distribution across different clients. The affine shift alters the data characteristics, such as mean and variance, simulating a more realistic scenario where data collected by different nodes may have other underlying distributions. 

Another partition protocol based on attributes is the ~\emph{Hist-Dirichlet}~\cite{jimenez2024fedartml}. The algorithm starts by characterizing the attributes of the centralized client (i.e., calculating the average of all the features) and categorizing it through a binning process. Next, the distribution of each feature class within each client is determined using the DD with a specified $\alpha$ applied to the binned variable. The latter ensures that the client data is divided into distinct, non-overlapping datasets.

\paragraph{\emph{Cluster-based}} This method uses clustering algorithms to group similar data points and distribute these clusters across clients, creating natural attribute differences.

The ~\emph{clustering} method~\cite{tang2024personalized} groups the dataset into distinct clusters based on inherent data similarities. These clusters are formed following a non-IID distribution, meaning that the data within each cluster may differ significantly in terms of its features. Once the clustering process is complete, the dataset is divided into datasets corresponding to the clusters, which is helpful for classification or regression tasks. This method ensures that each subgroup reflects a unique data distribution.

\paragraph{\emph{Noise-based}} Given an initial data federation, it is not genuinely a partition protocol but a method to add attribute differences among clients. It adds noise to the features of data points to create attribute skew, often using Gaussian noise~\cite{jimenez2024fedartml,nguyen2023federated,li2022federated,rizk2022federated,selialia2022federated} with varying parameters for different clients.

\subsubsection{\textbf{Quantity skew}} It deals with generating uneven distribution of data volume across clients. It is divided into two subcategories: Client distribution-based and Sample distribution-based.

\paragraph{\emph{Client distribution-based}} This approach allocates varying amounts of data to clients based on specified distributions or rules.

In the \emph{Dirichlet for quantity skew method}~\cite{li2022federated}, the size of the local datasets $|\mathcal{D}_i|$ varies across clients, even though the data distribution among them may remain consistent. A DD allocates different data samples to each client. Specifically, it samples $q \sim \text{Dir}_K(\beta)$, where each $q_j$ represents the proportion of total data assigned to client $C_j$. The parameter $\beta$ controls the degree of quantity imbalance. The latter method has one limitation: it is not usable when the data is too small or the number of clients is too big.

The ~\emph{Min-Size-Dirichlet} method~\cite{jimenez2024fedartml} helps partition data based on the client's data size. It corrects the limitation of the Dirichlet for quantity skew method. The algorithm starts by setting an $\alpha$ value for the DD to generate the participation proportions for each client. A minimum required number of samples is then established for each client, with the minimum proportion size defined as $MinSize = \frac{MinRequiredSize}{n}$, where $n$ represents the total number of samples in the centralized dataset. If any calculated proportions are smaller than $MinSize$, they are replaced by $MinSize$. Finally, the proportions are normalized between 0 and 1. The latter ensures the method will converge even when the dataset is small, or the number of clients is significant.

Beutel et.al.~\cite{beutel2020flower} implemented inside Flower datasets some \emph{power law partition} methods, including linear, exponential, and square partitions, to generate splits based on the partition ID. In the ~\emph{linear} partitioner, partition sizes increase linearly with the ID, so client 1 gets 1 sample, client 2 gets two samples, and client $k$ gets $k$ samples. The ~\emph{exponential} partitioner correlates partition sizes with the exponent of the ID, where client 1 gets $\lfloor e \rfloor$ units, client 2 gets $\lfloor e^2 \rfloor$, and so on, with leftover data from rounding added to the largest partition. In the ~\emph{square} partitioner, partition sizes increase quadratically, so client 1 gets 1 unit, client 2 gets four units, and client $k$ gets $k^2$ units.

\paragraph{\emph{Sample distribution-based}} This method creates quantity skew by sampling data points for each client according to certain probability distributions or step functions.

In the \emph{step partition} method~\cite{bao2024boba,chen2020fedbe}, each client is assigned a mixture of minor and significant classes, typically comprising more minor classes with relatively smaller datasets and fewer major classes containing larger datasets. The  $\alpha$ parameter regulates the ratio between the sizes of the major and minor class datasets. As the value of $\alpha$ increases, the degree of non-IIDness also escalates, leading to a more significant disparity in the data distribution among the clients.

From Fig.~\ref{fig:taxonomy_protocols}, we notice that \hlmycolor{partition protocols simulating and controlling two or three types of data skew simultaneously do not appear in the papers reviewed}. This observation reveals a significant gap in current partitioning approaches, as real-world FL scenarios present combinations of different skew types concurrently.

\subsection{Non-IIDness quantification metrics}




The term ”non-IID metric” is used to quantify how much a dataset deviates from being IID, measuring the degree of data heterogeneity. \hlmycolor{Although research on quantifying non-IIDness in FL is limited, its importance cannot be overstated} for several relevant reasons. First, it provides a clear understanding of the statistical diversity among participating clients, directly affecting model convergence, stability, and overall performance~\cite{guo2022fdqm}. Accurate measurements for measuring non-IIDness help researchers and practitioners anticipate challenges in training, allowing for the design of tailored algorithms to address the specific level of heterogeneity present. These metrics also enable standardized evaluations and benchmarking of the robustness of FL methods across different datasets and non-IID scenarios~\cite{haller2023handling}. Furthermore, they guide the development of standardized testbeds that resemble real-world FL scenarios more accurately to produce robust and generalizable FL solutions. 

\begin{figure}[ht] 
    \centering
    \includegraphics[width=\columnwidth]{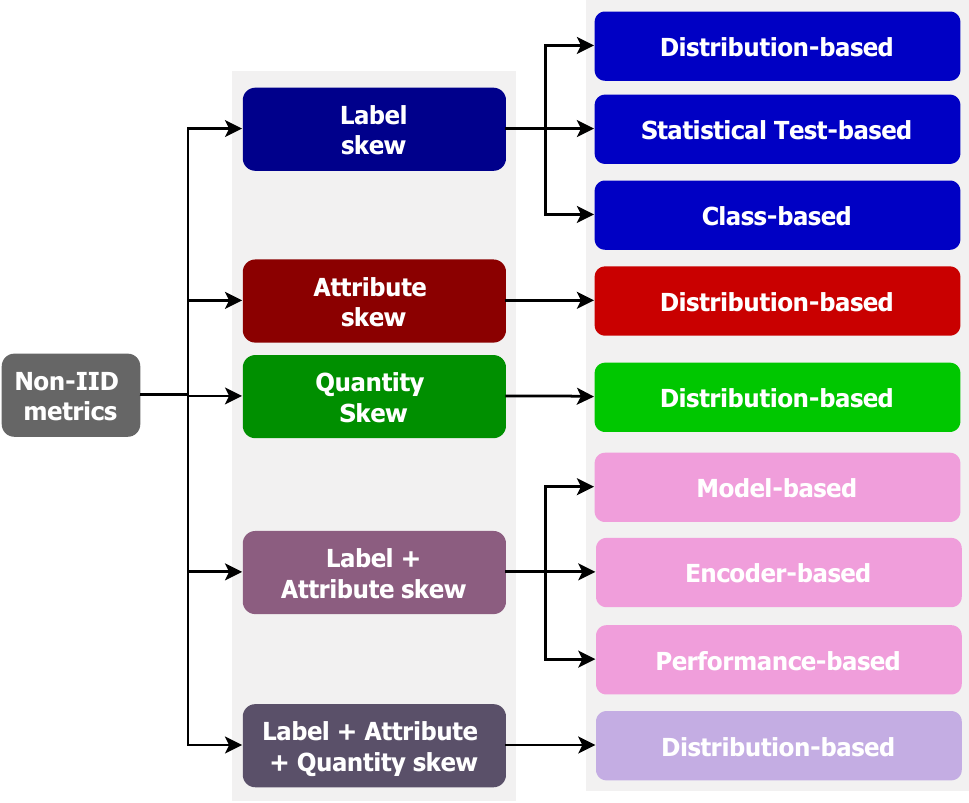}
    \caption{Taxonomy for non-IID metrics in Federated Learning}
    \label{fig:taxonomy_metrics}
\end{figure}

Based on the comprehensive review of papers selected in our survey, we developed a classification for non-IID metrics in recent literature. Fig.~\ref{fig:taxonomy_metrics} illustrates our taxonomy of measures used to quantify non-IID characteristics in FL research.

In the subsequent subsections, we provide definitions for each taxonomy category shown in Fig.~\ref{fig:taxonomy_metrics}. Additionally, we explain the specific metrics used to assess data heterogeneity within each category.

\subsubsection{\textbf{Label skew}} These metrics quantify the imbalance in class distributions across clients. This category is divided into Distribution-based, Statistical Test-based, and Class-based.

\paragraph{\emph{Distribution-based}} These metrics compare label distributions between clients or against a global distribution. Label distributions refer to the proportion of different classes within a dataset. The comparison uses widely known distances and divergences like Earth Mover's distance (also known as Wasserstein distance)~\cite{zhao2023non,chen2022emd,zhang2021federated,hsu2020federated}, Hellinger distance~\cite{jimenez2024fedartml,tan2023privacy}, Jensen-Shannon divergence~\cite{ahmed2023semisupervised,xufblg}, Kullback-Leibler divergence~\cite{zhang2023spatial}, and similarities like the Cosine~\cite{zhang2024addressing,shu2022clustered} and Jaccard~\cite{luo2024privacy}.

Despite their utility, the previous metrics exhibit limitations when applied to non-IID data distributions in FL. First, distance-based metrics like Earth mover's distance (Wasserstein) and Hellinger Distance, while effective in capturing dissimilarities, can become computationally expensive, especially in high-dimensional settings~\cite{ling2007efficient}. Furthermore, divergences such as Jensen-Shannon and Kullback-Leibler can be sensitive to small sample sizes, which may lead to inaccurate results when comparing distributions with limited data among clients. Lastly, similarity-based metrics like Cosine only consider the angle between two vectors, ignoring their magnitudes. The latter might result in misleading comparisons of label distributions between clients, especially when the data is highly imbalanced~\cite{srivastava2023new}.

\paragraph{\emph{Statistical Test-based}} These metrics leverage statistical tests to assess the significance of label distribution differences. It includes widely known tests such as the Kolmogorov-Smirnov statistic~\cite{qu2022rethinking} and the Chi-squared test~\cite{zawad2021curse}. Another alternative test is the Maximum Mean Discrepancy. The latter compares distributions by identifying a continuous function $f$, calculating its mean value for each distribution, and measuring the discrepancy between these means. This discrepancy reflects the distributions' differences, with larger discrepancies indicating greater differences.

\paragraph{\emph{Class-based}} These metrics focus on quantifying the non-IID ness based on the volume of participation of one or more classes compared to the maximum number of classes per client. Considering that fewer classes per client mean uneven data distribution and large heterogeneity of the dataset, Zawad et al.~\cite{zawad2021curse} proposed the \emph{Heterogeneity Index} ($HI$) as defined in Eq.~\ref{eq:HI}: 

\begin{equation} \label{eq:HI}
\begin{aligned} 
     HI = 1 - \frac{1}{(C_{\text{max}} - 1)} \cdot (c - 1)
\end{aligned}
\end{equation}

where $c$ is the maximum number of classes per client, and $C_{\text{max}}$ is the total number of classes in the dataset. Notice that $HI$ may oversimplify the complexity of data distributions, failing to capture nuanced variations in client data~\cite{park2023fedgeo}.

The \emph{Imbalance Ratio} ($IR$)~\cite{wen2023communication,ortigosa2017measuring} is another metric proposed in the literature defined as in Eq.~\ref{eq:IR}.

\begin{equation} \label{eq:IR}
\begin{aligned} 
     IR(\xi) = \frac{\max_i \xi_i}{\min_j \xi_j}
\end{aligned}
\end{equation}

where $\xi_i$ represents the frequency of class $i$ in the dataset. This metric measures the imbalance between the majority and minority classes. A higher IR value indicates a greater class imbalance in the dataset. The $IR$ has two main limitations: it is primarily suited for binary classification problems and may oversimplify multi-class imbalances by ignoring intermediate class distributions.

\subsubsection{\textbf{Attribute skew}} These metrics measure the differences in feature distributions across clients. This category includes only metrics using the attributes distribution per client (\emph{Distribution-based}). It includes measures like Hellinger distance, Earth mover's distance, and Jensen-Shannon divergence, applied to feature spaces rather than label spaces~\cite{jimenez2024fedartml}.

\subsubsection{\textbf{Quantity skew}} The metrics evaluate the imbalance in data volume across clients. This category includes only metrics leveraged on the client's participation distribution (\emph{Distribution-based}). It often uses measures similar to label and attribute skew (e.g., Hellinger distance, Earth mover's distance, Jensen-Shannon divergence) but applied to sample counts or proportions~\cite{jimenez2024fedartml}.

\subsubsection{\textbf{Label + Attribute skew}} \hlmycolor{These metrics combine label and attribute skew aspects to provide a more comprehensive assessment of data heterogeneity}. This category includes Model-based, Encoder-based, and Model Performance-based.

\paragraph{{\emph{Model-based}}} In this category, the metrics use the model to infer the combined effect of label and attribute skew. The ~\emph{model traveling} metric~\cite{hsieh2020non} estimates how well a model generalizes across different, often skewed, non-IID data partitions. During training, the model is periodically moved between different data partitions to assess its accuracy on other clients. Comparing the model's performance on its original data partition with its accuracy on a new partition allows for the estimation of accuracy loss, which reflects the degree of non-IIDness.

The other two metrics proposed in the literature are \emph{inner variation} and \emph{outer variation}~\cite{bao2024boba}. \emph{Inner variation} measures the randomness in sampling data from a client's local distribution $\FY{i}$ and is defined as  $\mathbb{E}[g_i] - \mathbb{E}[g_i^2]$, where $g_i$ is the gradient of client $i$. \emph{Outer variation} captures the difference between a client’s local distribution $P_i$ and the global distribution $\frac{1}{|H|} \sum_{i \in H} P_i$, and is given by $\mathbb{E}[g_i] - \mathbb{E}[u^2]$, where $u$ represents the global gradient. In IID settings, outer variation is zero, while in feature and label-skewed scenarios, it becomes non-zero, indicating higher non-IIDness.

\emph{Dataset entropy}~\cite{aamer2021entropy} is a metric designed to characterize a dataset's distribution, information quantity, unbalanced structure, and non-IID nature, independently of the models used. It is computed through a generalized clustering strategy, utilizing a custom similarity matrix that integrates features and supervised outputs, making it suitable for classification and regression tasks. The metric's reliance on clustering and a custom similarity matrix can lead to high computational overhead, potentially diminishing its cost-saving benefits in large FL setups.

\emph{The universal bound}~\cite{yang2021achieving} $s_G$ is a metric that indicates the extent to which local data distributions differ. A value of  $s_G = 0$ signifies that the dataset IID, while higher values indicate increasing levels of non-IID characteristics, reflecting more significant variability among the workers' data. This bound is crucial for analyzing the convergence rates of algorithms like $FedAvg$, as it provides insights into how non-IID conditions impact the training process and overall performance of FL models.

\paragraph{\emph{Encoder-based}} This type of metric uses learned representations from the data to quantify non-IIDness across both label and attribute dimensions. The only metric in this category \emph{The Client-Wise Non-IID Index (CNI)}~\cite{li2020lotteryfl}. The CNI measures how different the data distribution of a client $i$ is from that of other clients. It is defined as:

\begin{equation} \label{eq:CNI}
\resizebox{0.8\columnwidth}{!}{$
\text{CNI}(i) = \frac{\left\| \left( \frac{1}{|C_i|} \sum_{k} \text{En}\left(\mathcal{D}_i^k\right) \right) - \left( \frac{1}{|C_j|} \sum_{l \neq i} \text{En}\left(\mathcal{D}_j^l\right) \right) \right\|_2}{\sigma\left(\text{En}(\mathcal{D})\right)}
$}
\end{equation}

where \( \mathcal{D} = \bigcup_{i=1}^K \mathcal{D}_i \), \( \mathcal{D}_i^k \) denotes the data belonging to the \(k\)-th class in \( \mathcal{D}_i \). \( |C_i| \) is the number of classes in \( \mathcal{D}_i \), \( \sigma(\cdot) \) is the standard deviation and is used to normalize the scale, and \( \|\cdot\|_2 \) indicates the \( \ell_2 \)-norm. The intuition behind Eq.~\ref{eq:CNI} is to measure the distance between the average data representations from different classes in feature space on a given client and the counterpart over all the other clients.

\paragraph{\emph{Performance-based}} It refers to metrics that directly measure the impact of combined label and attribute skew on model performance. Under this category the only metric found is ~\emph{Dataskew}~\cite{haller2023handling}, defined as in Eq.~\ref{eq:DATASKEW}:

\begin{equation} \label{eq:DATASKEW}
\begin{aligned} 
    \text{Dataskew} = \frac{\max(\Delta \text{Accuracy}_{\text{pairwise}})}{\frac{1}{K} \sum_{i=1}^{K} \text{Accuracy}_{i}}
\end{aligned}
\end{equation}

Where \( \max(\Delta \text{Accuracy}_{\text{pairwise}}) \) is the maximum pairwise deviation of accuracy between clients. \( \frac{1}{K} \sum_{i=1}^{K} \text{Accuracy}_{_i} \) is the average accuracy across all clients. A high Dataskew value (close to 1) indicates strong data heterogeneity. It might introduce bias into the calculation as it includes the client on which the initial model was trained. If that client has a significantly different data distribution, it could skew the interpretation of the metric.

\subsubsection{\textbf{Label + Attribute + Quantity skew}} This category represents the most comprehensive metrics for all three main types of non-IIDness. \hlmycolor{It is noticeable that quantifying more than one skewness simultaneously is paramount to understanding the complete behavior of non-IIDness in the FL system.}

\paragraph{\emph{Distribution-based}} Based on the label, attribute, and quantity distribution distributions, these metrics attempt to capture the combined effects of such types of skew in a single measure. The \emph{Four-Dimensional Quantitative Measure (FDQM)}~\cite{guo2022fdqm} measures heterogeneity across four key dimensions:

\begin{itemize}
    \item \emph{Noise Estimation for Feature Distribution Skew}: This dimension quantifies the noise or variability in feature distributions across different clients, assessing how features deviate from one another.
    \item \emph{Degree of Difference for Label Distribution Skew}: This measures the difference in class label distributions among clients, capturing the extent of label imbalance.
    \item \emph{Degree of Difference for Quantity Skew}: This dimension evaluates the variability in the amount of data across clients, identifying differences in dataset sizes.
    \item \emph{Feature Distribution Estimation for Feature Differency}: This estimates the difference in feature distributions across clients, focusing on the dissimilarity in feature values or characteristics.
\end{itemize}

\emph{FDQM} combines these four dimensions into a single comprehensive metric that helps quantify non-IIDness. FDQM involves multiple statistical measures, such as noise estimation, Dirichlet concentration parameters, and Gaussian Mixture Models. These calculations can become computationally intensive, mainly when applied to large-scale FL systems with many clients. Additionally, it does not have an upper bound that can help to understand "acceptable" levels of heterogeneity. The metric assumes that client data follows specific statistical distributions. If the client data does not align with these assumptions, the FDQM may provide inaccurate or misleading heterogeneity estimates. 

Our survey reveals a notable \hlmycolor{absence of metrics that can simultaneously measure combinations of data heterogeneities, such as label with quantity skew, attribute with quantity skew, or combinations with spatiotemporal heterogeneity, participation skew, and modality skew}. This measurement gap is particularly concerning, given that these heterogeneities co-exist in real-world FL scenarios. Without combined metrics, researchers and practitioners lack the tools to effectively quantify and understand how multiple types of data heterogeneity jointly impact model performance.

%% file: Sections/6.solutions.tex
\section{Non-IID solutions in FL} \label{sec:non-iid-solutions}



\begin{figure}[ht]
    \centering
    \includegraphics[width=\columnwidth]{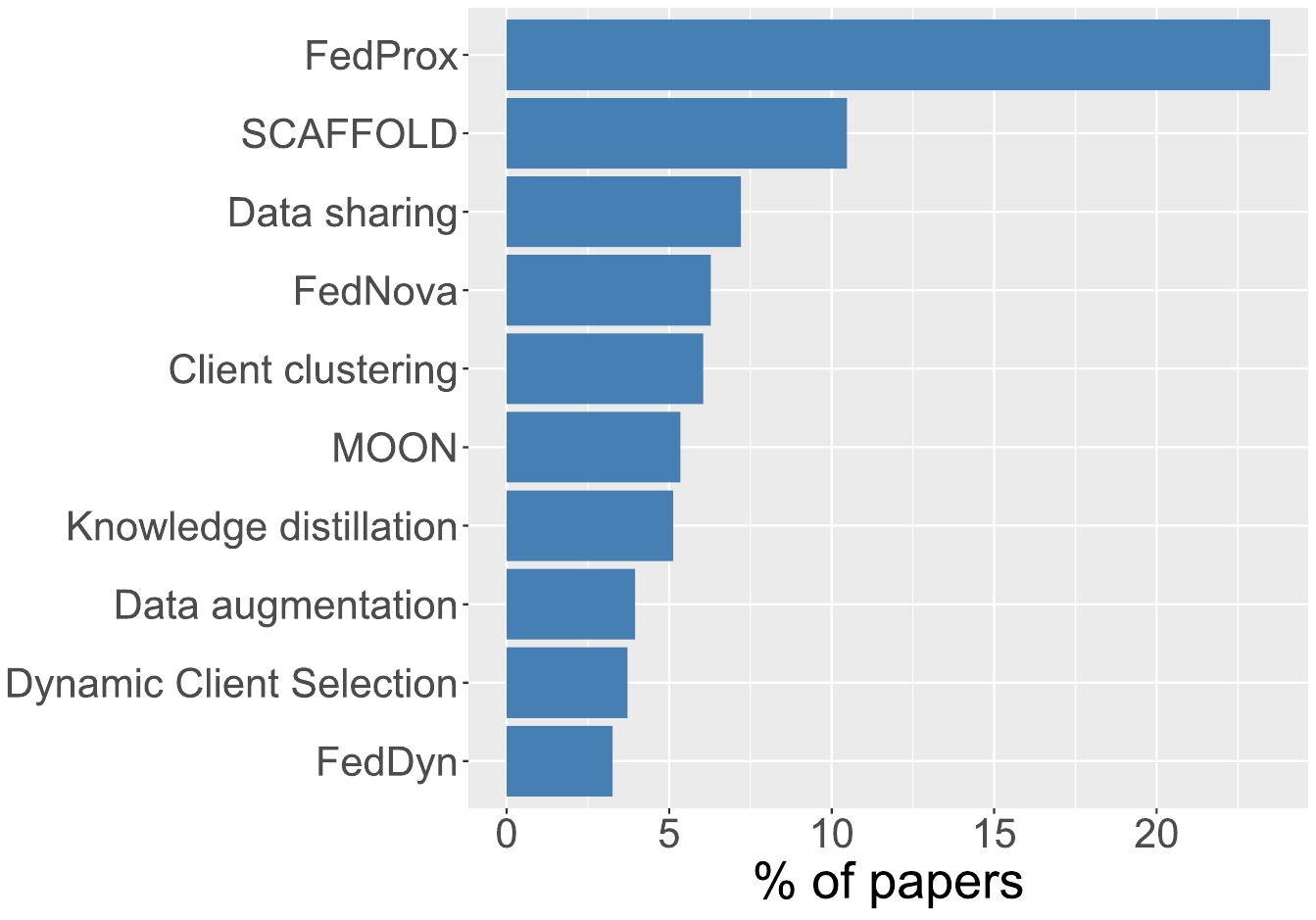}
    \caption{Relative prevalence of state-of-the-art solutions to tackle non-IIDness}
    \label{fig:top_alg_fed}
\end{figure}

This section briefly discusses the leading solutions for non-IID datasets in FL to tackle the downstream effects exposed in Section~\ref{sec:FL-basics}. \hlmycolor{Notice that most of these solutions focus on one type of non-IIDness, which is something future research should improve}. In particular, we describe in the following paragraphs the solutions that are employed the most in the selected papers of our survey (see Fig.~\ref{fig:top_alg_fed}).

\emph{FedProx~\cite{li2020federated}}. Designed to combat heterogeneity within FL environments, addressing both system and statistical heterogeneity. As an extension and reparameterization of the FedAvg method, FedProx introduces flexibility by enabling each participating device to perform varying amounts of work, adhering to individual device-level system constraints. The advantages of FedProx include providing convergence guarantees even when learning from non-IID data, thus mitigating the effects of statistical heterogeneity. Moreover, it offers the advantage of requiring only minor adjustments to existing methods, making it straightforward to implement and integrate into preexisting FL systems. However, it is worth noting that FedProx may still present some limitations. For example, its performance can be sensitive to the choice of the proximal term, and it may not fully address all forms of statistical heterogeneity, particularly in scenarios with extreme distribution differences.

\emph{Scaffold~\cite{karimireddy2020scaffold}}. It is an algorithm to mitigate the challenges posed by data heterogeneity among clients in an FL setting. This method leverages control variates, a variance reduction technique, to counteract the client drift phenomenon often occurring in local updates during the FL process. The critical advantage of Scaffold lies in its ability to address the performance issues stemming from non-IID client data. Through the strategic use of control variates, Scaffold effectively corrects the local update drift, enhancing FL's convergence and overall effectiveness. One potential constraint is the need for adequate domain knowledge and expertise in configuring the control variates, which may pose challenges in specific practical scenarios. 

\emph{Data sharing~\cite{zhao2018federated,seo2022resource,shao2022dres}}. Contrary to the standard FL approaches, where the server and the clients do not share data, this approach envisions that initially, the server trains a global model on a globally shared dataset. A fraction of this dataset is delivered to the clients that update their local models, employing the local private training data and the shared global data. Experiments show that accuracy can be increased by 30\% for the CIFAR-10 dataset with only 5\% globally shared data. Despite providing good performance, this method suffers from some shortcomings: it is difficult to get an initial good-quality global dataset since the server has no idea about the data distributions among the connected clients, and sharing data violates the requirement of privacy-preserving learning, which is a fundamental motivation of FL.


\emph{FedNova~\cite{wang2020tackling}}. FedNova is a normalized averaging method that targets resolving the objective inconsistency issue commonly encountered in FL. It eliminates objective inconsistency while maintaining rapid error convergence, making it a valuable addition to the FL landscape. The advantages of FedNova include its ability to provide a systematic understanding of the solution bias and convergence slowdown resulting from objective inconsistency in FL. While it addresses objective inconsistency, its performance and applicability can still be influenced by the specific characteristics of the FL problem, including the degree of heterogeneity among the clients' data distributions.

\emph{Client clustering~\cite{cong2024ada,tun2023contrastive,lee2023fedlc,ma2023flgan,zou2022efficient}}.
Most FL approaches assume that the whole system distills only one global model. However, this may be a challenging and unrealistic assumption, especially in heterogeneous data environments. In client clustering, this assumption is relaxed. Instead of training a single FL model among all clients, clients form clusters characterized by more balanced and homogeneous data and then perform FL within the individual clusters. This approach results in more performant intra-cluster models. 

The clusters should be created without disclosing sensitive client information to the central server to preserve privacy. To this purpose, the literature proposes two main methods: the similarity of the loss values and the similarity of model weights. In the former approach, the server constructs multiple global models instead of a single one and shares them with the clients. Then, the clients can compute a local empirical loss and update the received cluster model with the smallest loss value. Finally, the updated cluster model is returned to the server for cluster model aggregation. In the latter approach, an initial global model is downloaded to each device to update the weights locally, which are returned to the server. The server can finally derive the similarity scores employing received model weights and group the clients into clusters given such scores~\cite{tian2022wscc}. 

\emph{Model-contrastive learning (MOON)~\cite{li2021model}}. It uses the similarity between model representations to correct the local training of individual parties, i.e., conducting contrastive learning at the model level. The key idea of contrastive learning is to reduce the distance between the representations of different augmented views of the same image (i.e., positive pairs) and increase the distance between the representations of augmented views of different images (i.e., negative pairs). Specifically, MOON corrects the local updates by maximizing the agreement of representation learned by the current local model and the representation learned by the global model. Extensive experiments show that MOON significantly outperforms the other state-of-the-art FL algorithms on various image classification tasks. 

\emph{Knowledge Distillation (KD)~\cite{wang2024novel,madni2024robust,chen2023privacy,zhang2023memory,he2022class}}.
It is a technique where a larger, more complex model, known as the "teacher," transfers its knowledge compactly and efficiently to smaller, simpler models, referred to as the "student." The compact and efficient way to transfer knowledge is soft labels, which represent the probabilities the teacher model assigns to each class. The latter conveys not just the final decision but also the confidence levels across potential outcomes. The goal is to transfer the comprehensive knowledge of the teacher model to a student model that retains much of the teacher's accuracy and performance but with significantly fewer parameters. This typical teacher-student framework in KD is quite suitable for FL. The idea is that well-performing models (teachers) can transfer their knowledge to poorly performing ones (students). KD-based FL algorithms only require clients to exchange their local models' soft labels (e.g., logits) without uploading model parameters or data, thus significantly reducing potential privacy risks and communication costs. 

\emph{Data Augmentation~\cite{zhang2024secure,guo2023new,zhao2023ensemble,li2022federated3}}. It has been employed to increase the diversity of training data by applying suitable (e.g., random but realistic) transformations to the original dataset. The same technique can mitigate local data imbalance issues in FL using three main data augmentation methods: vanilla, mixup, and generative adversarial network (GAN).
In vanilla data augmentation, each client sends its label distribution information to the server. The server computes the global statistics, considering the information from each client, and informs the client on the degree of augmentation needed to mitigate the imbalance. The original local data sample and the augmented data are used together to update the local model parameters.

The core idea of mixup is to synthesize new samples by combining existing data points. To preserve privacy, the existing data points are encoded and delivered to the server that performs the mixup operation. The obtained balanced dataset is used to train a global model that is then delivered to the clients until the training converges.

Federated GANs are designed to train a generator in the presence of non-IID data. The generator is trained in the server, employing the data shared by the clients, and it is finally delivered to the clients. They can eventually use it to replenish the training data for the missing components (e.g., labels) to obtain a more balanced local dataset.
Despite being proven effective, most of these techniques rely on data sharing, which might increase the risk of data privacy leakage.


\emph{Dynamic Client Selection~\cite{zhang2024addressing,cho2022towards,chen2022emd,pandey2022contribution,zhang2021adaptive,cho2020client}}. Most FL papers assume unbiased client participation, where clients are selected randomly or in proportion to their data sizes. In Dynamic Client Selection, clients are selected employing a bias strategy. In other words, clients are dynamically chosen at each round to achieve a good trade-off between convergence speed and solution bias. Its advantages include faster convergence by prioritizing clients with higher-quality data, better handling system heterogeneity, and efficient use of resources. However, it may introduce selection bias, leading to over-reliance on specific clients, increased bias, and decreased model generalization. Additionally, it can increase the complexity of client selection, add communication overhead, and risk underrepresented clients with limited computational resources.

\emph{FedDyn~\cite{jin2023feddyn}} This method introduces a novel approach to construct a proxy dataset and extract local knowledge dynamically. Instead of using the average strategy, they implemented focus distillation to emphasize reliable knowledge, effectively addressing the non-IID issue where local models may have biased knowledge. The average strategy weakens knowledge quality by treating reliable and unreliable knowledge equally. Additionally, they applied local differential privacy techniques on the client side to safeguard private user information from local knowledge. Experimental results demonstrated that their method achieves faster convergence and reduces communication overhead compared to baseline approaches.

In FL, the choice between personalized and a single global model significantly impacts how heterogeneity is addressed. A global model simplifies collaboration but struggles with performance when client data distributions differ significantly, often requiring advanced optimization techniques like FedProx or Scaffold to mitigate these effects. In contrast, personalized models (a.k.a personalized FL~\cite{tan2022towards}) allow each client to tailor the model to their local data, improving accuracy for clients with skewed or specialized datasets but adding complexity and computational cost~\cite{muhammad2021robust}. Techniques like client clustering, KD, or dynamic client selection aim to balance personalization with shared knowledge. The decision between personalized models and a global model ultimately depends on factors such as the degree of data heterogeneity across clients, the need for model consistency, the application’s tolerance for varying performance across clients, and the available computational resources for handling the added complexity of personalization~\cite{wang2024towards,nguyen2022enhancing}.

It is worth noting that Fig.~\ref{fig:top_alg_fed} illustrates the prevalence of each solution discussed in this section. From the papers retrieved, we can see that FedProx is the most commonly employed solution, accounting for nearly 25\% of the total participation. In contrast, solutions like Scaffold, Data Sharing, and FedNova are utilized, but their participation rates are significantly lower than those of FedProx.

%% file: Sections/7.Frameworks_for_FL.tex
\section{Frameworks and tools for heterogeneous data in FL}\label{sec:frameworks}

FL has witnessed the development of numerous standardized frameworks to simplify its implementation, including partition protocols to federate centralized data and tackle the challenges of non-IID scenarios. This section comprehensively reviews notable FL frameworks, emphasizing their distinctive characteristics for partitioning, quantifying,  and handling non-IID data. 


\emph{FEDML}~\cite{he2020fedml} Also known as TensorOpera AI, it offers a comprehensive FL research and deployment platform, supporting various partition protocols and non-IID solutions. \emph{Flower}~\cite{beutel2020flower} provides a flexible framework for FL, emphasizing partition protocols in their Flower Datasets library. \emph{TensorFlow Federated (TFF)}~\cite{TFF} integrates FL capabilities into the TensorFlow ecosystem, offering tools for simulating simple FL scenarios. \emph{PySyft}~\cite{ziller2021pysyft}, while limited in partition protocols, focuses on privacy-preserving ML techniques. \emph{OpenFL}~\cite{foley2022openfl} aims to standardize FL workflows across different industries, particularly healthcare and finance. \emph{FLUTE}~\cite{hipolito2022flute} specializes in FL for natural language processing tasks, addressing unique challenges in text data. \emph{Federated AI Technology Enabler (FATE)}~\cite{liu2021fate} provides a comprehensive suite for secure FL, emphasizing privacy and security in collaborative AI. Lastly, \emph{FedLab}~\cite{zeng2023fedlab} supports various partition protocols, offering a robust environment for researching heterogeneous data scenarios in FL.

In addition to standardized frameworks, tools and benchmarks have been developed to facilitate research and evaluation in FL with non-IID data. \emph{FedArtML}~\cite{jimenez2024fedartml} is a comprehensive toolkit explicitly designed for exploring and analyzing the impact of non-IID data in FL scenarios, offering a range of partition protocols and non-IID metrics. \emph{LEAF}~\cite{caldas2018leaf} provides a modular benchmarking framework for learning in federated settings, featuring a collection of open-source datasets and a suite of evaluation tools to assess FL algorithms under various data distribution scenarios. NIID-Bench \emph{NIID-Bench}~\cite{li_federated_2022} provides a standardized benchmarking platform, providing researchers with a consistent environment to compare different approaches and assess their effectiveness in handling data heterogeneity.

Table~\ref{tab:frameworks_compare} comprehensively compares various standardized frameworks and tools, focusing on their capabilities in partitioning centralized data into non-IID federated data and tackling non-IID issues. The comparison is divided into three main categories: Partition protocols, non-IID metrics, and non-IID solutions.
Regarding partition protocols, FedLab, FedArtML, and LEAF appear to be the most comprehensive, supporting label, attribute, and quantity skewness. Most frameworks support label skew, but attribute skew is less commonly supported. Notably, PySyft seems to lack support for any partition protocols.

\begin{table*}[t!]
  \resizebox{\textwidth}{!}{\begin{tabular}{lccccccccccc}
    \toprule
     & \textbf{FEDML} & \textbf{Flower} & \textbf{TFF} & \textbf{PySyft} & \textbf{OpenFL} & \textbf{FLUTE} & \textbf{FATE} & \textbf{FedLab} & \textbf{FedArtML} & \textbf{LEAF} & \textbf{NIID-Bench}\\
    \midrule
    \textbf{Partition protocols} & \textcolor{orange}{\faMinusCircle} & \textcolor{teal}{\faCheckCircle} & \textcolor{orange}{\faMinusCircle} & \textcolor{purple}{\faTimesCircle} & \textcolor{orange}{\faMinusCircle} & \textcolor{orange}{\faMinusCircle} & \textcolor{orange}{\faMinusCircle} & \textcolor{teal}{\faCheckCircle} & \textcolor{teal}{\faCheckCircle} & \textcolor{teal}{\faCheckCircle} & \textcolor{orange}{\faMinusCircle} \\
    \midrule
    \hspace{2mm}\ding{72} Label skew & \textcolor{orange}{\faMinusCircle} & \textcolor{teal}{\faCheckCircle} & \textcolor{orange}{\faMinusCircle} & \textcolor{purple}{\faTimesCircle} & \textcolor{orange}{\faMinusCircle} & \textcolor{teal}{\faCheckCircle} & \textcolor{orange}{\faMinusCircle} & \textcolor{teal}{\faCheckCircle} & \textcolor{teal}{\faCheckCircle} & \textcolor{teal}{\faCheckCircle} & \textcolor{teal}{\faCheckCircle} \\
    \hspace{2mm}\ding{72} Attribute skew & \textcolor{purple}{\faTimesCircle} & \textcolor{purple}{\faTimesCircle} & \textcolor{purple}{\faTimesCircle} & \textcolor{purple}{\faTimesCircle} & \textcolor{purple}{\faTimesCircle} & \textcolor{purple}{\faTimesCircle} & \textcolor{purple}{\faTimesCircle} & \textcolor{teal}{\faCheckCircle} & \textcolor{teal}{\faCheckCircle} & \textcolor{teal}{\faCheckCircle} & \textcolor{orange}{\faMinusCircle} \\
    \hspace{2mm}\ding{72} Quantity skew & \textcolor{orange}{\faMinusCircle} & \textcolor{teal}{\faCheckCircle} & \textcolor{purple}{\faTimesCircle} & \textcolor{purple}{\faTimesCircle} & \textcolor{purple}{\faTimesCircle} & \textcolor{purple}{\faTimesCircle} & \textcolor{purple}{\faTimesCircle} & \textcolor{teal}{\faCheckCircle} & \textcolor{teal}{\faCheckCircle} & \textcolor{orange}{\faMinusCircle} & \textcolor{orange}{\faMinusCircle} \\    
    \midrule
    \textbf{non-IID metrics} & \textcolor{purple}{\faTimesCircle} & \textcolor{orange}{\faMinusCircle} & \textcolor{purple}{\faTimesCircle} & \textcolor{purple}{\faTimesCircle} & \textcolor{purple}{\faTimesCircle} & \textcolor{purple}{\faTimesCircle} & \textcolor{purple}{\faTimesCircle} & \textcolor{purple}{\faTimesCircle} & \textcolor{orange}{\faMinusCircle} & \textcolor{purple}{\faTimesCircle}  & \textcolor{purple}{\faTimesCircle} \\
    \midrule
    \hspace{2mm}\ding{72} Label skew & \textcolor{purple}{\faTimesCircle} & \textcolor{orange}{\faMinusCircle} & \textcolor{purple}{\faTimesCircle} & \textcolor{purple}{\faTimesCircle} & \textcolor{purple}{\faTimesCircle} & \textcolor{purple}{\faTimesCircle} & \textcolor{purple}{\faTimesCircle} & \textcolor{purple}{\faTimesCircle} & \textcolor{teal}{\faCheckCircle} & \textcolor{purple}{\faTimesCircle} & \textcolor{purple}{\faTimesCircle} \\
    \hspace{2mm}\ding{72} Attribute skew & \textcolor{purple}{\faTimesCircle} & \textcolor{purple}{\faTimesCircle} & \textcolor{purple}{\faTimesCircle} & \textcolor{purple}{\faTimesCircle} & \textcolor{purple}{\faTimesCircle} & \textcolor{purple}{\faTimesCircle} & \textcolor{purple}{\faTimesCircle} & \textcolor{purple}{\faTimesCircle} & \textcolor{teal}{\faCheckCircle} & \textcolor{purple}{\faTimesCircle}  & \textcolor{purple}{\faTimesCircle} \\
    \hspace{2mm}\ding{72} Quantity skew & \textcolor{purple}{\faTimesCircle} & \textcolor{purple}{\faTimesCircle} & \textcolor{purple}{\faTimesCircle} & \textcolor{purple}{\faTimesCircle} & \textcolor{purple}{\faTimesCircle} & \textcolor{purple}{\faTimesCircle} & \textcolor{purple}{\faTimesCircle} & \textcolor{purple}{\faTimesCircle} & \textcolor{teal}{\faCheckCircle} & \textcolor{purple}{\faTimesCircle} & \textcolor{purple}{\faTimesCircle} \\
    \hspace{2mm}\ding{72} Label + Attribute skew & \textcolor{purple}{\faTimesCircle} & \textcolor{purple}{\faTimesCircle} & \textcolor{purple}{\faTimesCircle} & \textcolor{purple}{\faTimesCircle} & \textcolor{purple}{\faTimesCircle} & \textcolor{purple}{\faTimesCircle} & \textcolor{purple}{\faTimesCircle} & \textcolor{purple}{\faTimesCircle} & \textcolor{purple}{\faTimesCircle} & \textcolor{purple}{\faTimesCircle}  & \textcolor{purple}{\faTimesCircle} \\
    \hspace{2mm}\ding{72} Label + Attribute + Quantity skew & \textcolor{purple}{\faTimesCircle} & \textcolor{purple}{\faTimesCircle} & \textcolor{purple}{\faTimesCircle} & \textcolor{purple}{\faTimesCircle} & \textcolor{purple}{\faTimesCircle} & \textcolor{purple}{\faTimesCircle} & \textcolor{purple}{\faTimesCircle} & \textcolor{purple}{\faTimesCircle} & \textcolor{purple}{\faTimesCircle} & \textcolor{purple}{\faTimesCircle} & \textcolor{purple}{\faTimesCircle} \\ 
    \midrule
    \textbf{non-IID solutions} & \textcolor{teal}{\faCheckCircle} & \textcolor{teal}{\faCheckCircle} & \textcolor{purple}{\faTimesCircle} & \textcolor{orange}{\faMinusCircle} & \textcolor{purple}{\faTimesCircle} & \textcolor{orange}{\faMinusCircle} & \textcolor{purple}{\faTimesCircle} & \textcolor{teal}{\faCheckCircle} & \faMinus & \faMinus  & \faMinus  \\
    \midrule
    \hspace{2mm}\ding{72} FedProx & \textcolor{teal}{\faCheckCircle} & \textcolor{teal}{\faCheckCircle} & \textcolor{teal}{\faCheckCircle} & \textcolor{teal}{\faCheckCircle} & \textcolor{teal}{\faCheckCircle} & \textcolor{teal}{\faCheckCircle} & \textcolor{teal}{\faCheckCircle} & \textcolor{teal}{\faCheckCircle} & \faMinus & \faMinus  & \faMinus  \\
    \hspace{2mm}\ding{72} SCAFFOLD & \textcolor{teal}{\faCheckCircle} & \textcolor{teal}{\faCheckCircle} & \textcolor{purple}{\faTimesCircle} & \textcolor{teal}{\faCheckCircle} & \textcolor{purple}{\faTimesCircle} & \textcolor{teal}{\faCheckCircle} & \textcolor{purple}{\faTimesCircle} & \textcolor{teal}{\faCheckCircle} & \faMinus & \faMinus  & \faMinus  \\
    \hspace{2mm}\ding{72} Data sharing & \textcolor{purple}{\faTimesCircle} & \textcolor{orange}{\faMinusCircle} & \textcolor{purple}{\faTimesCircle} & \textcolor{purple}{\faTimesCircle} & \textcolor{purple}{\faTimesCircle} & \textcolor{purple}{\faTimesCircle} & \textcolor{purple}{\faTimesCircle} & \textcolor{purple}{\faTimesCircle} & \faMinus & \faMinus  & \faMinus  \\
    \hspace{2mm}\ding{72} FedNova & \textcolor{teal}{\faCheckCircle} & \textcolor{teal}{\faCheckCircle} & \textcolor{purple}{\faTimesCircle} & \textcolor{purple}{\faTimesCircle} & \textcolor{purple}{\faTimesCircle} & \textcolor{purple}{\faTimesCircle} & \textcolor{purple}{\faTimesCircle} & \textcolor{teal}{\faCheckCircle} & \faMinus & \faMinus  & \faMinus  \\
    \hspace{2mm}\ding{72} Client clustering & \textcolor{purple}{\faTimesCircle} & \textcolor{orange}{\faMinusCircle} & \textcolor{purple}{\faTimesCircle} & \textcolor{purple}{\faTimesCircle} & \textcolor{purple}{\faTimesCircle} & \textcolor{purple}{\faTimesCircle} & \textcolor{purple}{\faTimesCircle} & \textcolor{teal}{\faCheckCircle} & \faMinus & \faMinus  & \faMinus  \\
    \hspace{2mm}\ding{72} MOON & \textcolor{purple}{\faTimesCircle} & \textcolor{teal}{\faCheckCircle} & \textcolor{purple}{\faTimesCircle} & \textcolor{purple}{\faTimesCircle} & \textcolor{purple}{\faTimesCircle} & \textcolor{purple}{\faTimesCircle} & \textcolor{purple}{\faTimesCircle} & \textcolor{purple}{\faTimesCircle} & \faMinus & \faMinus  & \faMinus  \\
    \hspace{2mm}\ding{72} Knowledge distillation & \textcolor{teal}{\faCheckCircle} & \textcolor{teal}{\faCheckCircle} & \textcolor{purple}{\faTimesCircle} & \textcolor{purple}{\faTimesCircle} & \textcolor{purple}{\faTimesCircle} & \textcolor{purple}{\faTimesCircle} & \textcolor{purple}{\faTimesCircle} & \textcolor{purple}{\faTimesCircle} & \faMinus & \faMinus  & \faMinus  \\
    \hspace{2mm}\ding{72} Data augmentation & \textcolor{purple}{\faTimesCircle} & \textcolor{purple}{\faTimesCircle} & \textcolor{purple}{\faTimesCircle} & \textcolor{purple}{\faTimesCircle} & \textcolor{purple}{\faTimesCircle} & \textcolor{purple}{\faTimesCircle} & \textcolor{purple}{\faTimesCircle} & \textcolor{purple}{\faTimesCircle} & \faMinus & \faMinus  & \faMinus  \\
    \hspace{2mm}\ding{72} Dynamic client selection & \textcolor{purple}{\faTimesCircle} & \textcolor{orange}{\faMinusCircle} & \textcolor{purple}{\faTimesCircle} & \textcolor{purple}{\faTimesCircle} & \textcolor{purple}{\faTimesCircle} & \textcolor{purple}{\faTimesCircle} & \textcolor{purple}{\faTimesCircle} & \textcolor{orange}{\faMinusCircle} & \faMinus & \faMinus  & \faMinus  \\
    \hspace{2mm}\ding{72} FedDyn & \textcolor{teal}{\faCheckCircle} & \textcolor{purple}{\faTimesCircle} & \textcolor{purple}{\faTimesCircle} & \textcolor{purple}{\faTimesCircle} & \textcolor{purple}{\faTimesCircle} & \textcolor{purple}{\faTimesCircle} & \textcolor{purple}{\faTimesCircle} & \textcolor{teal}{\faCheckCircle} & \faMinus & \faMinus & \faMinus  \\

    \bottomrule
  \end{tabular}}
    \caption{FL standardized frameworks and tools for non-IID data comparison (\textcolor{teal}{\faCheckCircle}: Sufficient, \textcolor{orange}{\faMinusCircle}: Under development/incomplete, \textcolor{purple}{\faTimesCircle}: Unknown/Not implemented), \faMinus: Not applicable}
  \label{tab:frameworks_compare}

\end{table*}

Regarding non-IID metrics, FedArtML is the only framework with comprehensive support for label, attribute, and quantity skew metrics. Most frameworks have limited or no support for non-IID metrics, and no framework fully supports metrics for combined skews (label + attribute or label + attribute + quantity). For non-IID solutions, FEDML and Flower appear to be the most versatile. FedProx is universally supported across all applicable frameworks, while more advanced solutions like MOON, data augmentation, and dynamic client selection are less commonly supported. There is a general lack of support for non-IID metrics and combined skew scenarios across most frameworks. 

\begin{figure}[ht]
  \centering
  \includegraphics[width=\linewidth]{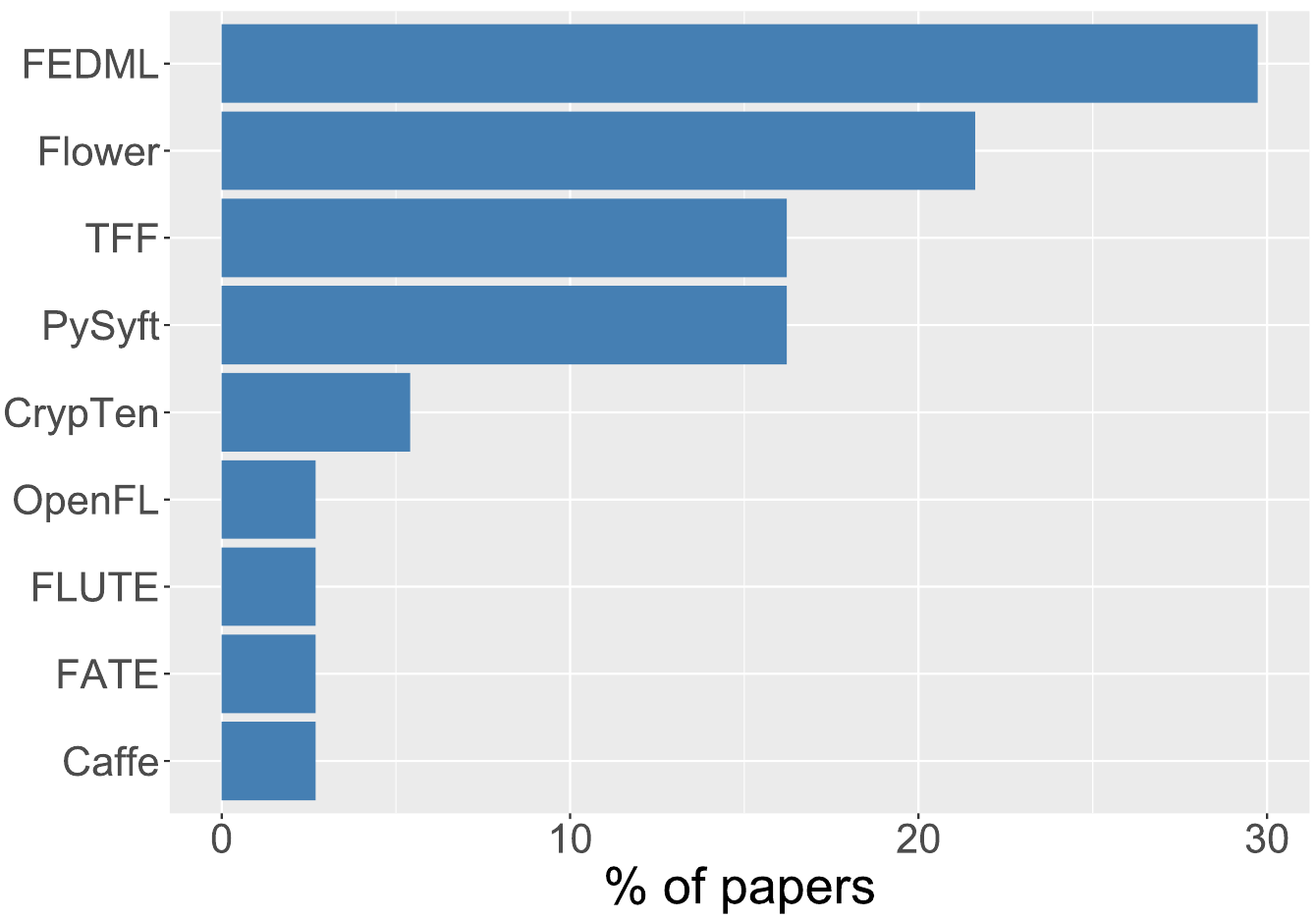}
  \caption{Relative prevalence of use for standardized frameworks in FL}
  \label{fig:frameworks_particip}
\end{figure}

Fig.~\ref{fig:frameworks_particip} illustrates the adoption rates of standardized frameworks for FL across the research papers studied. It reveals FEDML as the clear leader, utilized in approximately 30\% of papers. Flower is the second most popular framework, appearing in about 20\% of studies. TFF and PySyft show similar usage rates, each featuring in roughly 18-19\% of papers. Lastly, OpenFL, FLUTE, and FATE exhibit more limited use, accounting for less than 5\% of the articles.

While reviewing the literature retrieved, \hlmycolor{we realized that, surprisingly, \emph{only 14.2\% of them use FL frameworks to implement their proposals}}. This low adoption rate suggests that a substantial portion of the research in FL is conducted in isolation, potentially relying on custom or ad-hoc implementations rather than established frameworks designed to facilitate FL experiments. The latter disconnect may hinder the reproducibility and comparability of results, as custom implementations can vary widely in terms of functionality and efficiency. Encouraging greater integration of FL frameworks could lead to more consistent experimental practices, foster collaboration, and accelerate advancements by leveraging these established tools' robust features and optimizations.

%% file: Sections/8.Lessons.tex
\section{Essential findings and key insights}\label{sec:lessons}




This section highlights the essential findings and critical insights obtained through our survey of the current literature on non-IID data in FL. We aim to provide practitioners with robust conclusions and thought-provoking considerations for developing practical solutions in this challenging area of FL. These lessons encompass various aspects, including the quantification and classification of non-IID-ness, the interplay between different types of data skew, and the practical implications for implementation and standardized framework selection. 

\begin{enumerate}[leftmargin=*]
    \item \textbf{Lack of consensus on non-IID classification and quantification}.
Researchers have no agreement on how to classify and quantify non-IID-ness in FL. This lack of standardization makes it challenging to compare different studies and solutions. Furthermore, no consensus exists on sufficient conditions or scenarios (regarding data partition and simulations) to test new solutions addressing non-IID-ness adequately.

    \item \textbf{Importance of quantifying non-IID-ness}. Among the studied papers, just 13.1\% of them employed metrics to quantify the level of non-IIDness~\cite{hu2024fedmmd,liao2024rethinking,shang2022fedic,ma2021fast,yang2021achieving}. Using such metrics to quantify non-IID-ness is crucial for adequately characterizing FL datasets. Without such quantification, assessing the severity of non-IID issues and the effectiveness of proposed solutions becomes difficult.

    \item \textbf{Interdependence of skew types}. When partitioning datasets to create a specific type of skew (e.g., label skew), researchers often inadvertently introduce side effects in other types of skewness.  Nevertheless, they do not take into account such effects during their experimentation. For example, in the research papers reviewed, 60.3\% of the papers that include label skew approaches do not mention anything about quantity skew, even if the partition methods employed generate the mentioned side effect~\cite{chen2024fed,wang2024delta,wang2024dfrd,lee2023can,zheng2023federated,elbatel2023federated,wang2023distribution,luo2022fedsld,yu2022spatl,maeng2022towards}. This interdependence highlights the complexity of non-IID scenarios and the need for a more holistic approach to dataset preparation and analysis.

    \item \textbf{Prevalence of label skew studies}. Our survey indicates that a majority of papers focus on label skew. While this is an important aspect of non-IID data, causing the major deterioration of FL models~\cite{morafah2023practical,ou2022aggenhance,he2022improving,abdel2022privacy,majeed2022comparative,chiu2020semisupervised}, the disproportionate attention may lead to overlooking other critical types of non-IID skews.

    \item \textbf{Limitations of single skew analysis}. Quantifying only one skew type at a time is insufficient for fully understanding and addressing non-IID-ness in FL. In real-life scenarios, the data is generally affected by more than one type of skewness, and the simulations performed should mimic such settings. Thus, a more comprehensive approach that simultaneously considers multiple types of skewness is necessary to develop robust solutions.

    \item \textbf{Custom FL implementations vs. standardized frameworks}. Only 14.2\% of the reviewed papers employed standardized FL frameworks (i.e., FEDML, Flower, Pysyft, etc.) to implement their solutions to tackle non-IIDness. While this approach offers flexibility, it also introduces risks such as implementation errors and reduced reproducibility. 

    \item \textbf{Few works in multimodality skew research}. The multimodality skew, as exposed in Section~\ref{sec:data-heterogeneity}, is a topic that has not been formally included in the previous surveys for non-IIDness. Despite its relevance, the research papers that include that topic are just a few~\cite{chen2024fed,borazjani2024multi,yuan2024communication,tan2023fedsea,feng2023fedmultimodal,gao2022new,chen2022towards,chai2020fedat}. As real-world FL applications often involve diverse data types and modalities, understanding and addressing multimodality skew is a key aspect for practical implementations with the aim of understanding and tackling the consequences of non-IID data in FL.

\end{enumerate}

%% file: Sections/9.Future_directions.tex
\section{Future directions and trends}\label{sec:future}
This section delineates promising future research directions and emerging trends for developing innovative solutions to address non-IID data challenges in FL. We aim to highlight key areas that can guide researchers in formulating novel approaches and methodologies to enhance our understanding of the impact of non-IIDness on FL systems. 

\begin{enumerate}[leftmargin=*]
    \item \textbf{Evaluation and benchmarking}. A critical area for future research in non-IID FL is the development of more sophisticated evaluation and benchmarking methodologies. Real-life FL data is typically affected by multiple types of skewness simultaneously, a nuance often overlooked in current simulations~\cite{wang2024analyzing,shen2024decentralized,khan2023precision,huang2023rethinking,jeong2023tutorial,jiang2022harmofl}. Future research should focus on developing comprehensive partition protocols that consider various forms of data heterogeneity concurrently, such as label skew, feature skew, and quantity skew simultaneously. This enhanced evaluation framework will lead to more reliable performance assessments and drive the creation of FL algorithms that are more resilient and adaptable to diverse non-IID scenarios.
    
    \item \textbf{Theoretical advancements}. Theoretical advancements represent a crucial frontier in addressing non-IID challenges in FL. A primary focus should be the development of new, more sophisticated metrics for quantifying data heterogeneity. Current measures often fail to capture the full complexity of non-IID scenarios, particularly when multiple types of data skewness coexist. Future research should aim to create comprehensive metrics that can accurately reflect the multidimensional nature of data heterogeneity, incorporating as many data skews as possible.
    
    Additionally, significant effort should be directed toward improving generalization bounds for non-IID FL. The development of tighter bounds on generalization performance is essential for bridging the gap between theoretical guarantees and practical performance. The latter refers to theoretical guarantees or limits on how well a model trained on a distributed dataset can perform on unseen data~\cite{gholami2024improved,wei2022non,ro2021communication,mohri2019agnostic}. These improved bounds should account for various aspects of data heterogeneity and federation strategies, providing more accurate model performance predictions across diverse non-IID settings. Such theoretical advancements will leverage the development of more robust and efficient algorithms capable of maintaining high performance in severe data heterogeneity.
    
    \item \textbf{Modality skew research}. A promising avenue for future research in FL is the exploration of multimodal learning techniques to address the increasingly complex nature of real-world data. As applications of FL expand across diverse domains, there is a growing need to develop robust methods capable of handling multiple data modalities simultaneously, such as text, images, audio, and sensor data~\cite{yu2023multimodal,ouyang2023harmony}. Future research should focus on creating novel FL algorithms that can effectively integrate and learn from these heterogeneous data types while preserving privacy and maintaining efficiency. This direction presents unique challenges in the context of non-IID data, as different modalities may exhibit varying degrees and types of heterogeneity across clients. 
    
    Therefore, it is vital to develop new metrics to accurately quantify multimodality skew, capturing the distribution differences within each modality and the relationships and dependencies between different modalities. These metrics should be capable of measuring aspects such as cross-modal correlation discrepancies, modality-specific feature disparities, and variations in the relative importance of different modalities across clients. The latter helps to get the full potential of diverse, real-world datasets while effectively managing the complexities of non-IID data distributions.
    
    \item \textbf{Addressing data heterogeneity impact}. Addressing data heterogeneity remains a central challenge in FL, necessitating innovative approaches to enhance model performance and adaptability in non-IID settings~\cite{shang2023evolutionary,chen2023fraug,shuai2022balancefl}. Future research should prioritize the development of more effective algorithms designed to handle diverse forms of non-IIDness, specifically under high levels of non-IIDness. These advanced algorithms should aim to mitigate the negative impacts of non-IID data on model convergence, bias, and generalization (robustness). 
    
    Additionally, there is a pressing need for adaptive methods capable of dynamically adjusting to different non-IID scenarios encountered in real-world applications. Such methods could leverage online learning techniques or meta-learning approaches to rapidly identify and adapt to varying degrees and types of data heterogeneity across clients or over time. Furthermore, the emerging field of Federated Neural Architecture Search (FNAS) presents a promising direction for automatically designing optimal model architectures in federated settings, particularly under non-IID conditions~\cite{yao2024perfedrlnas,yan2024peaches,liu2023finch,zhu2021federated2}. FNAS could potentially uncover novel network structures that are inherently more robust to data heterogeneity, leading to improved performance and efficiency in diverse FL scenarios. 
    
    \item \textbf{Communication efficiency}. Enhancing communication efficiency remains another critical area for future research in FL. As the scale and complexity of federated systems grow, the need for optimized communication protocols becomes increasingly paramount. Future research should focus on designing advanced compression techniques tailored to model updates in non-IID environments~\cite{silvi2024accelerating,arisdakessian2023coalitional,gao2023fedios,cao2023knowledge,chen2022contractible,li2021hermes}. These techniques must strike a delicate balance between reducing communication overhead and preserving the integrity of model updates, which is especially challenging when clients have disparate data distributions. Innovative approaches might include adaptive compression methods that adjust their strategies based on the degree of data heterogeneity or the importance of specific model parameters. 
    
    Exploring asynchronous communication protocols presents another promising avenue for improving efficiency and scalability in FL systems. Asynchronous methods could allow for more flexible participation of clients, potentially mitigating some of the challenges posed by non-IID data distributions~\cite{zakerinia2024communication,xu2023asynchronous}. However, careful consideration must be given to ensuring model convergence and managing the potential staleness of updates in asynchronous settings. The latter enables the deployment of FL in a broader range of real-world scenarios with varying network conditions and client capabilities.
    
    \item \textbf{Model generalization and task adaptation}. A final area for future research in FL is enhancing model generalization and task adaptation in non-IID data. As real-world data distributions often evolve, there is a pressing need to investigate methods to improve model robustness and adaptability to these dynamic scenarios~\cite{tan2023taming,yang2023federated,huang2023train,chen2023prompt,zhu2022diurnal,li2022federated2,sery2021over}. Future work should focus on developing techniques to mitigate catastrophic forgetting in non-IID settings, where new data distributions may cause the model to lose performance on previously learned tasks rapidly. The latter could involve exploring continual learning approaches or developing novel regularization techniques specifically designed for federated environments with heterogeneous and time-varying data. 
    
    While much of the current research in FL has centered on image classification tasks, there is a significant opportunity to extend these methods to a broader range of domains and applications. Future research should aim to adapt and optimize FL algorithms for diverse tasks such as natural language processing, speech recognition, and time series analysis~\cite{yang2024fedfed,zhang2022fine,wang2022non,yang2021characterizing,li2021fedrs}. This expansion will require addressing the challenges posed by non-IID data in these domains, including handling variable-length inputs, dealing with domain-specific data skew, and managing the increased complexity of model architectures often required for these tasks. 

    \item \textbf{Public datasets with non-IIDness quantification}. A crucial step toward advancing research on non-IID data in FL is the creation and dissemination of publicly available datasets with explicit and standardized characterizations of their non-IID properties. Including clear non-IID metrics to quantify the extent and types of non-IIDness in these datasets would facilitate benchmarking, reproducibility, and the development of more robust methods tailored to diverse real-world scenarios.
\end{enumerate}

%% file: Sections/10.Conclusion.tex
\section{Conclusion}\label{sec:conclusions}


In summary, this work offers a comprehensive technical survey of the state-of-the-art non-IID data in FL. Unlike existing surveys, we include a detailed taxonomy for non-IID data, partition protocols and metrics to quantify non-IIDness, popular solutions to address data heterogeneity, and standardized frameworks employed in FL with heterogeneous data. We include modality skew as a novel category in the taxonomy data heterogeneity, which appears as a new trend among the researchers. We indicate a lack of consensus on non-IID classification and quantification, highlighting the importance of using metrics to measure non-IIDness. We claim that quantifying only one skew type at a time is insufficient for fully understanding and addressing non-IIDness. Finally, we suggest some research directions that can be adopted in future investigations in the field.